\documentclass[10pt]{article} 
\usepackage[accepted]{rlc}

\usepackage{microtype}
\usepackage{graphicx}
\usepackage{subfigure}
\usepackage{booktabs} 
\usepackage{svg}
\usepackage{subcaption}

\usepackage{stackengine}
\usepackage{hyperref}

\usepackage{amssymb}            
\usepackage{mathtools}          
\usepackage{mathrsfs}           
\usepackage{graphicx}           
\usepackage{subcaption}         
\usepackage[space]{grffile}     
\usepackage{url}                
\usepackage{algorithm}
\usepackage{algorithmic}

\usepackage{amsmath}
\usepackage{amssymb}
\usepackage{mathtools}
\usepackage{amsthm}
\DeclareMathOperator*{\argmin}{arg\,min}

\usepackage[capitalize,noabbrev]{cleveref}
\usepackage{wrapfig}

\newcommand\blfootnote[1]{%
  \begingroup
  \renewcommand\thefootnote{}\footnote{#1}%
  \addtocounter{footnote}{-1}%
  \endgroup
}

\theoremstyle{plain}
\newtheorem{theorem}{Theorem}[section]

\newtheorem{lemma}[theorem]{Lemma}

\theoremstyle{definition}
\newtheorem{definition}[theorem]{Definition}
\newtheorem{assumption}[theorem]{Assumption}
\theoremstyle{remark}

\title{Learning Action-based Representations Using \\ Invariance}

\author{Max Rudolph\(^{*1}\), Caleb Chuck\(^{*1}\), Kevin Black\(^{*2}\), Misha Lvovsky\(^{3}\),\\\textbf{Scott Niekum\(^{3}\), Amy Zhang\(^{1}\)}\\
The University of Texas at Austin\(^{1}\) \\
University of California, Berkeley\(^{2}\) \\
University of Massachusetts Amherst\(^{3}\)  \\
}

\begin{document}

\maketitle
\blfootnote{* Authors contributed equally, corresponding author \texttt{mrudolph@cs.utexas.edu}}
\begin{abstract}
Robust reinforcement learning agents using high-dimensional observations must be able to identify relevant state features amidst many exogeneous distractors. 
A representation that captures \textit{controllability} identifies these state elements by determining what affects agent control. While methods such as inverse dynamics and mutual information capture controllability for a limited number of timesteps, capturing long-horizon elements remains a challenging problem. Myopic controllability can capture the moment right before an agent crashes into a wall, but not the control-relevance of the wall while the agent is still some distance away. To address this we introduce \textit{action-bisimulation encoding}, a method inspired by the bisimulation invariance pseudometric, that extends single-step controllability with a recursive invariance constraint.
By doing this, action-bisimulation learns a multi-step controllability metric that smoothly discounts distant state features that are relevant for control. We demonstrate that action-bisimulation pretraining on reward-free, uniformly random data improves sample efficiency in several environments, including a photorealistic 3D simulation domain, Habitat.
Additionally, we provide theoretical analysis and qualitative results demonstrating the information captured by action-bisimulation. Code and video: \href{https://maxrudolph1.github.io/action-bisimulation-site/}{https://maxrudolph1.github.io/action-bisimulation-site/}
\end{abstract}



\section{Introduction}


Learning control for complex decision-making from high-dimensional observation spaces such as video and depth is vital for real-world applications of reinforcement learning (RL). To do this, a representation of the observation space allows agents to reason about the environment and take intelligent actions. However, learning these representations is often sample inefficient. One reason for this is that real-world scenarios often contain many irrelevant and distracting features embedded in a high-dimensional space. Correlating reward with relevant state elements, and not causally confusing distractors in this setting, is challenging---especially since reward signals are often sparse.

Representation learning has emerged as a promising approach to address this challenge by extracting a compressed and informative representation of the observation space that is useful for learning~\citep{bengio2013representation}. Representation learning removes irrelevant distractors from the state space used to learn the policy, which improves sample efficiency and performance. In RL, task-specific representation learning uses reward or expert behavioral similarity~\citep{ferns2011bisimulation,agarwal2021contrastive} to discover the compressed representation, only describing task-specific elements. This has the advantage of capturing only information that is either useful for solving the task or relevant to the demonstrations while being limited by requiring either expert behavior or task-achieving policies, both of which can be difficult to obtain prior to learning. On the other hand, task-free methods use unsupervised signals like reconstruction~\citep{lange2010deep} and contrastive objectives~\citep{laskin2020curl} and can be pre-trained on any data, including random actions. However, these methods are trained without action information. As a result they can capture exogenous distractors that are not useful for improving RL policy performance.

One promising direction of task-agnostic methods utilizes \textit{controllability} to learn a behavior-relevant representation that is not task-specific~\citep{lambguaranteed}. These representations can avoid capturing task-irrelevant information while not requiring expert or reward-achieving behavior. Recent work in action-based representation learning for RL has shown promising results \citep{zhang2022learning} by utilizing inverse dynamics models to extract representations \citep{islam2022agent}. These representations rely on a window of information by predicting the first action between two states separated by $k$-steps. If $k$ is small this representation is myopic, but when $k$ is large the prediction problem is underspecified. This underspecification restricts large $k$ to offline datasets with correlated action data---such as expert trajectories.

\begin{figure}
\centering     

\subfigure[]{\includegraphics[width=0.42\textwidth]{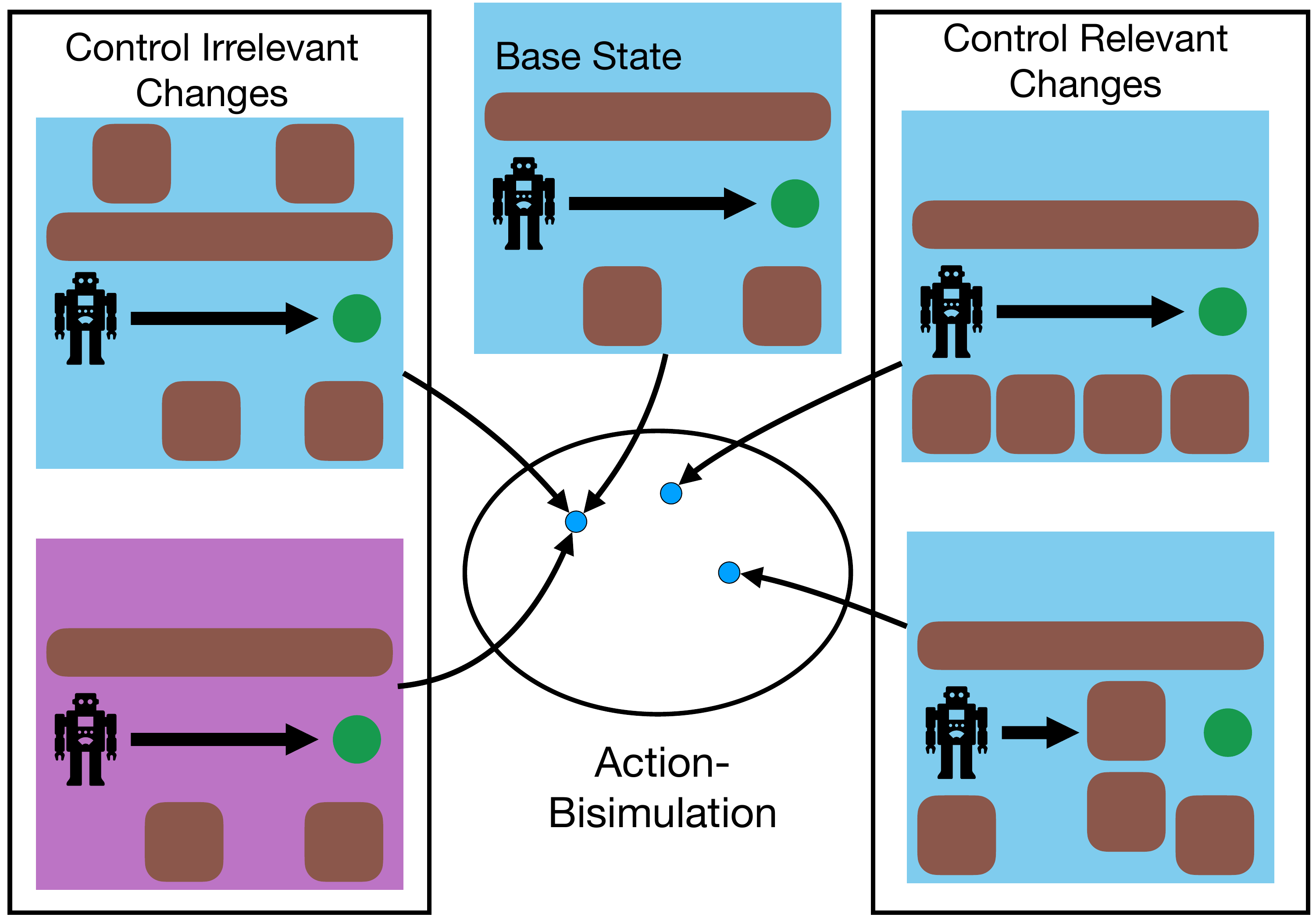}}
\hspace{1cm}
\subfigure[]{\includegraphics[width=0.5\textwidth]{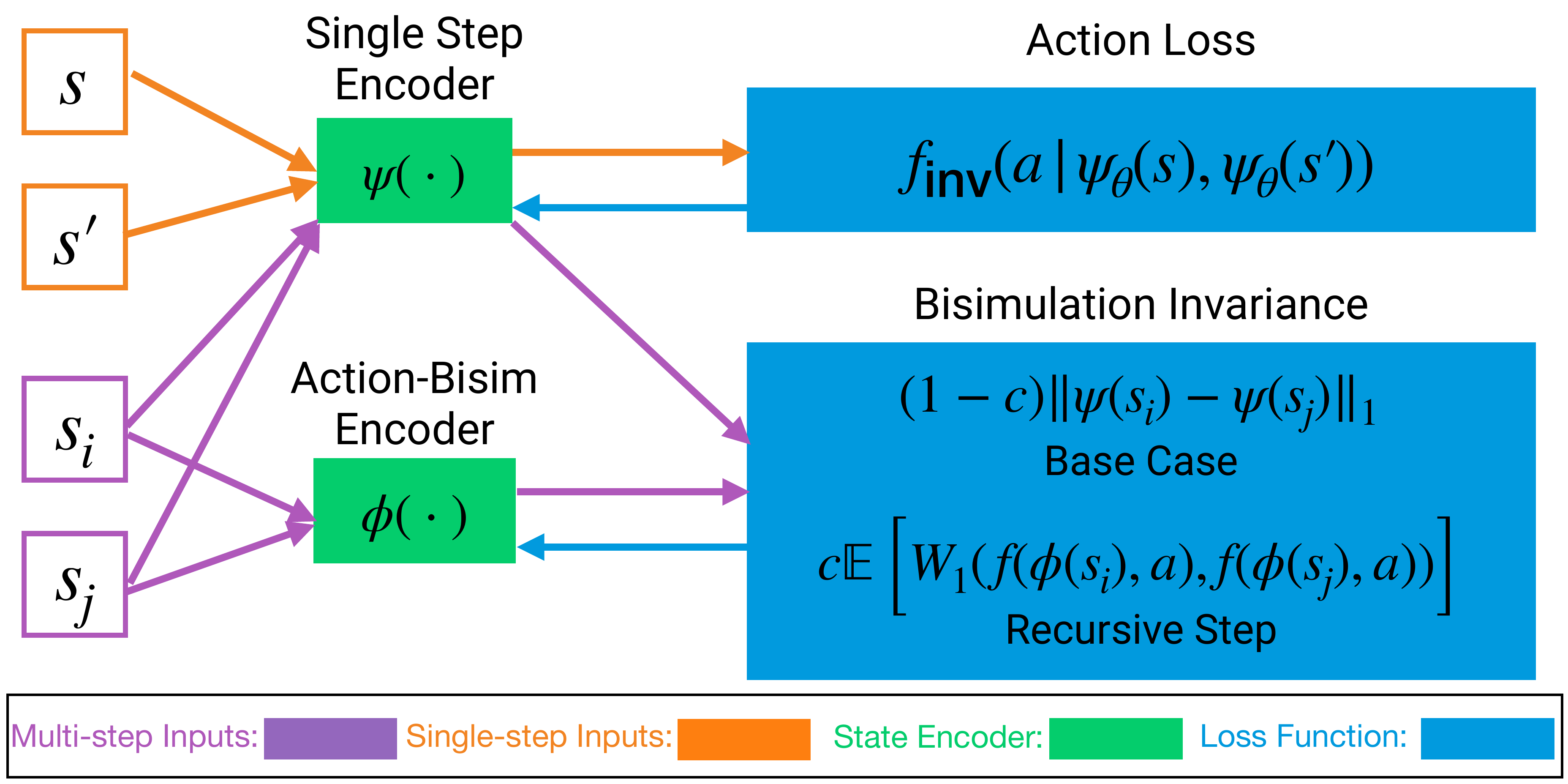}}
\caption{\textbf{Left: mapping equivalent} \textit{controllability} together with action-bisimulation. Other methods can be too aggressive (single-step inverse dynamics would map together \textbf{(a)} and \textbf{(e)}, reward-based methods would map together \textbf{(a)} and \textbf{(d)}), or permissive (autoregressive methods would map \textbf{(a)} and \textbf{(c)} to the same value). \textbf{Right: data flow} of action-bisimulation training. The single-step encoder is trained with inverse dynamics (Section~\ref{sec:controllability_measures}). The multi-step encoder is trained with bootstrapped single-step representation distance (Equation~\ref{actionbisimulation}).}
\vskip -0.5cm
\label{fig:bisim_main}
\end{figure}

We investigate utilizing a novel invariant metric to learn a multi-step control-based representation instead of directly applying $k$-step prediction. Our \textit{action-bisimulation metric} offers a novel framework for controllability metrics that takes a myopic dynamics encoding and extends it to multi-step representations. This formulation is inspired by reward bisimulation~\citep{zhang2020learning}, which utilizes single-step reward information to learn multi-step return-capturing representations. Action-bisimulation applies bootstrapping on the myopic $k=1$ controllability representations to enforce multi-step invariance in an \textit{action-bisimulation encoding}. Since the base case uses single-step prediction, the encoding can be trained with any offline data, even fully random. At the same time, boostrapping extends the action-bisimulation encoding to capture long-term controllability. \cref{fig:bisim_main} captures how action-bisimulation maps control-irrelevant states together, while not doing the same for control-relevant states.

This work offers an empirical analysis and theoretical formulation of the novel control-based invariant metric for representation learning. We demonstrate empirically that in scenarios where complex, long-horizon, sparse-reward decision-making is required, the metric improves sample efficiency compared to RL agents trained directly from pixels, or pre-trained with existing representation learning methods in multiple domains. Next, we provide qualitative results demonstrating the robustness of the learned representation to uncontrollable 
distractors, as well as sensitivity to control-relevant state features.

\section{Related Works}
\vspace{-0.3cm}

\textbf{Representation Learning in RL}. Learned representations have been widely applied to RL, formalized~\citep{li2006towards} through hierarchical symbolic representations~\citep{konidaris2014constructing,andre2002state}, skill abstractions~\citep{dietterich2000hierarchical}, policy optimality~\citep{auer2008near,jong2005state,abel2016near}, selective attention~\citep{jones2010integrating} and contingency awareness~\citep{bellemare2012investigating}. One effective strategy is to use the representation to learn a model that is effective for planning~\citep{hafner2019dream,koul2023pclast}. These methods learn world models~\citep{ha2018world} and other representations that can be used for prediction~\citep{singh2012predictive}, data generation and planning. Alternatively, other methods apply representation learning for filtering~\citep{krishnan2015deep, karl2016deep} or reduced complexity 
~\citep{higgins2016beta,oord2018representation,laskin2020curl} representations. 
Action-bisimulation is a novel encoder that learns controllability-based representations to improve RL performance. Unlike other representation learning methods, action-bisimulation uses a soft invariance pseudometric to capture action information through time.

\textbf{Action-based Representations}. RL methods have directly leveraged action-relevant representations in several ways. This includes contingency awareness~\citep{bellemare2012investigating,choi2018contingency,chuck2020hypothesis,chuck2023granger}, which is closely related to action controllability~\citep{zhong2020disentangling} and control information measures like empowerment (channel capacity between actions and state)~\citep{jung2011empowerment,mohamed2015variational,levy2023hierarchical} or affordances~\citep{cruz2016training,khetarpal2020can,nagarajan2020ego}. Multi-step inverse models are most similar to action-bisimulation, but common multi-step inverse methods~\citep{lambguaranteed, islam2022agent, koul2023pclast} require selecting a specific $k$ for the multi-step horizon, potentially leaving critical control information on the table. Further, it has been shown that multi-step inverse models can be insufficient when the dynamics are periodic~\citep{levine2024multistep}. Action-bisimulation uses a soft invariance metric to extend single-step models, which better preserves long-term controllability.

\textbf{Bisimulation methods}. Bisimulation describes future invariant state representations, originally applied to stationary representations~\citep{larsen1989bisimulation,dean1997model,ferns2004Metrics}, before being extended to continuous state MDPs~\citep{ferns2011bisimulation}. Reward-based bisimulation methods have gained popularity through learned deep representations~\citep{zhang2020learning}.
This has been extended to non-optimal policies~\citep{castro2021mico}, with generalized value function bounds~\citep{kemertas2021towards} and augmented with state discretization~\citep{kemertas2022approximate} and clustering~\citep{liu2023robust}. Bisimulation-based methods have also been applied in different contexts: expert policy similarity~\citep{agarwal2021contrastive,bertran2022efficient,mazoure2021cross}, goal-conditioned RL~\citep{hansen2022bisimulation} and reward-action policy equivalence~\citep{liao2023policy,castro2020scalable}. While this work draws on reward-bisimulation, action-bisimulation is fundamentally offline and task-agnostic because it takes an expectation over actions, removing its dependence on any policy.  

\vspace{-0.2cm}
\section{Preliminaries}
\label{preliminaries}
\vspace{-0.2cm}
A Markov decision process is defined by the tuple $\mathcal M \coloneqq (\mathcal S, \mathcal A, p, R)$, where $\mathcal S$ is the state space, $\mathcal A$ is the action space and $s\in \mathcal S, a \in \mathcal A$ are states and actions respectively. $p(s'|s,a)$ is the transition function that gives the probability of the next state $s'$ given the current state and action $(s,a)$. The reward function $R(s,a)$ maps state and action to a scalar reward. A policy $\pi(a|s)$ is the probability of an action given the current state.

This work utilizes the following two-phase paradigm: in the first phase, the agent first takes actions without access to the reward function $R(s,a)$ to generate a dataset of ordered state action tuples $\mathcal D \coloneqq \{(s^{(0)},a^{(0)}), \hdots(s^{(|\mathcal D|-1)},a^{(|\mathcal D|-1)})\}$. Then, a representation $\phi: \mathcal S \rightarrow \mathcal Z$ is learned from $\mathcal S$. In the second phase, the agent learns from extrinsic reward utilizing the learned representation. 

The action-bisimulation representation method is inspired by reward bisimulation~\citep{dean1997model}. In RL, bisimulation is a state abstraction that groups reward-equivalent states:
\begin{definition}[Bisimulation Relations~\citep{givan2003equivalence}]
    \label{defn-bisimulation}
 In MDP $\mathcal M$, an equivalence relation $B$ between states is a bisimulation relation if: $\forall s_i, s_j \in \mathcal S$ where the states are equivalent under $B$ ($s_i \equiv_B s_j$), the following
conditions hold:
\begin{align}
R(s_i, a) &= R(s_j,a) \quad \forall a \in \mathcal A \\
P(\mathcal G|s_i,a) &= P(\mathcal G|s_j,a)\quad \forall a \in \mathcal A, \forall \mathcal G \in \mathcal S_B
\end{align}
where $\mathcal S_B$ is the partition of $\mathcal S$ under the relation $B$ (the set of all groups 
$\mathcal G$ of equivalent states), and $P(\mathcal G|s, a) = \sum_{s' \in \mathcal G}p(s'|s,a)$
\end{definition}

Bisimulation Metrics~\citep{ferns2011bisimulation,castro2020scalable} soften the notion of state partitions with a pseudometric space $(\mathcal S, d)$, where distance function $d:\mathcal S \times \mathcal S \rightarrow R_{\geq 0}$ measures the similarity between two states.\footnote{This is a pseudometric, meaning that two different states can have $0$ distance.} The on-policy bisimulation metric~\citep{kemertas2021towards} is:
\begin{equation}
d_\text{r-bisim}(s_i, s_j) = \max_a\underbrace{(1-c)\cdot |R(s_i,a) - R(s_j, a)|}_{\text{base case}}  + \underbrace{cW_1(d)(p(\cdot|s_i, a), p(\cdot|s_j,a))}_\text{recursive step}, 
\label{reward_bisim}
\end{equation}


where $W_1(d)$ is the 1-Wasserstein distance and $c$ is a scalar hyperparameter that weights the multi-step sensitivity of the distance. The 1-Wasserstein metric measures the distance between next-state distributions in the latent bisimulation space.
We propose a novel controllability-based relation, which replaces reward equivalence with single-step control equivalence. By replacing rewards in the equivalence, the relation is task-agnostic.

\begin{definition}[Action-Bisimulation Relations]
    \label{defn-action-bisimulation}
 Let $\psi:\mathcal S \rightarrow \mathcal Z_{ss}$ be a single step controllability encoder such that $p(a|\psi(s), \psi(s')) = p(a|s,s')$ for all $s,a,s'$. In MDP $\mathcal M$, an equivalence relation $AB$ between states is an action-bisimulation relation according to $\psi$ if: $\forall s_i, s_j \in \mathcal S$ where the states are equivalent under $AB$ ($s_i \equiv_{AB} s_j$), the following
conditions hold:
\begin{align}
\psi(s_i) &= \psi(s_j) \\
P(\mathcal G|s_i,a) &= P(\mathcal G|s_j,a)\quad \forall a \in \mathcal A, \forall \mathcal G \in \mathcal S_{AB}
\end{align}
where $\mathcal S_{AB}$ is the partition of $\mathcal S$ under the relation $AB$ (the set of all groups 
$\mathcal G$ of equivalent states), and $P(\mathcal G|s, a) = \sum_{s' \in \mathcal G}p(s'|s,a)$
\end{definition}

This equivalence can be similarly relaxed into a pseudometric. However, in the off-policy setting, we are not interested in a particular policy, but all policies. Thus, action-bisimulation uses the expectation over uniform actions to encode all possible policies.
\begin{equation}
d_\text{a-bisim}(s_i, s_j, \psi) =  \underbrace{(1-c)\cdot\|\psi(s_i) - \psi(s_j)\|_1}_\text{base case} 
 + \underbrace{c \cdot \mathbb{E}_{a\sim U(\mathcal A)} \left[ W_1(p(\cdot|s_i, a), p(\cdot|s_j, a)) \right]}_\text{recursive step}
\label{actionbisimulation}
\end{equation}
In the next section, we describe how $\psi(\cdot)$ can be learned from data, and how to use $d_\text{a-bisim}$ to learn an action-bisimulation encoder.

\section{Methods}
This section describes the algorithm for training an action-bisimulation encoder. First, the single-step encoder is learned, then the distance in single step space is used as the ``base case'' for the recursive step. The training flow and inputs are visualized in~\cref{fig:bisim_main}b.
\subsection{Single-Step Controllability}
\label{sec:controllability_measures}

Inverse dynamics describes the probability of an action given two sequential states $(s, s')$: 
$P(a|s,s')$. To get a single-step encoding of the action-relevant state features
we define the single step state encoder $\psi_\theta(s): \mathcal S \rightarrow \mathcal Z_{ss}$, where $\mathcal Z_{ss}$ is the embedded single-step space~\citep{lambguaranteed}, and $\psi_\theta$ is parameterized by $\theta$. Then, for dataset $\mathcal D$ of $(s,a,s')$ tuples, the regularized single-step representation is learned by optimizing the single-step (ss) inverse dynamics loss:
\begin{equation}
L_{ss}(\mathcal D, \theta, \nu) = -\sum_{(s,a,s')\sim \mathcal D}\log f_{\nu,\text{inverse}}(a|\psi_\theta(s), \psi_\theta(s')) + \beta \left(\|\psi_\theta(s)\|_1 + \|\psi_\theta(s')\|_1\right),
\label{RegularizedInverseDynamics}
\end{equation}
where $f_{\nu,\text{inverse}}$ is a learned inverse dynamics model parameterized by $\nu$. The regularization ensures that the learned representation includes the minimum information necessary to capture the action-dependent inverse dynamics.

This inverse model is optimized to predict a distribution over actions $P(\cdot| \psi_\theta(s), \psi_\theta(s'))$ using the single-step embeddings as inputs. 
In this work, we represent the parameters of the distribution as a function of $[\psi_\theta(s), \psi_\theta(s')]$. 
Intuitively, $\psi(\cdot)$ embeds control-relevant features by embedding action-relevant components of the state. We use a relatively weak inverse model under the intuition that the simpler the model used to capture inverse dynamics, the more information is forced into the embedding rather than the inverse dynamics model.


\subsection{Action-Bisimulation Metric}
This section describes how the action-bisimulation metric (Equation~\ref{actionbisimulation}) is used to learn an encoder $\phi_\eta(s): \mathcal S \rightarrow \mathcal Z$, where $\mathcal Z$ is the representation space and $\phi_\eta$ is parameterized by $\eta$. This definition uses the single-step representation space $\mathcal Z_{ss}$ to define the multi-step representation space $\mathcal Z$.



The recursive step $\mathbb{E}[cW_1(d)(p(\cdot|s_i, a), p(\cdot|s_j,a))]$ requires computing $p(\cdot|s_i, a)$ and $p(\cdot|s_j,a)$. This can be done by learning a forward model parameterized by $\upsilon$: $f_\upsilon(\phi_\eta(s_i), a): \mathcal Z \times \mathcal A\rightarrow P(\cdot|\phi_\eta(s_i), a)$ that takes in the state embedding and action and outputs a probability distribution over the next embedded state. We model this by outputting the parameters of a conditional Gaussian model $\mathcal N (\mu, \Sigma)$ following the practice of~\cite{zhang2020learning}. Using the notation $f(\phi_\eta(s_i), a)[s']$ to denote the probability of state $s'$ under the distribution $f_\upsilon(\phi_\eta(s_i), a)$, we train the forward model by minimizing the negative log-likelihood of the observed data in $\mathcal D$: 
\begin{equation}
\begin{aligned}
L_\text{forward}(D) = -\sum_{s,a,s' \sim D}\log f_\upsilon(\phi_\eta(s), a)[\phi_\eta(s')].
\end{aligned}
\label{forward_model_loss}
\end{equation}
In deterministic dynamics, the 1-Wasserstein distance equals the $l_1$ distance of the mean. $f_\upsilon(\cdot)$ is a function of the encoded state $\phi(s)$ rather than the observation $s$ because forward dynamics over the observations is more costly due to the inherent reconstruction objectives they minimize; this reconstruction could bring in uncontrollable elements and does not inherently include control centric components.

In the off-policy setting, we propose using one of two expectations for the recursive step: over the uniform distribution of actions $E_{a\sim U(\mathcal A)}$ or over the behavior distribution: $E_{a\sim \pi_b(s_i)}$. The use of the behavioral distribution applies to settings where random actions might restrict the distribution of observed states. In practice, these are computed using the empirical mean.
Then, the action-controllability bisimulation metric using learned models is:
\begin{equation}
d_\text{a-bisim}(s_i, s_j, \psi_\theta, \phi_\eta) =  (1-c)\cdot\|\psi_\theta(s_i) - \psi_\theta(s_j)\|_1 
 + c \cdot \mathbb{E}_{a\sim U(\mathcal A)} \left[ W_1(f(\phi_\eta(s_i), a), f(\phi_\eta(s_j), a)) \right]
\label{actionbisimulationlearn}
\end{equation}

To train the encoder, we match the $l_1$ distance between the embedded representations $\phi(s_i), \phi(s_j)$ to the metric distance:
\begin{equation}
L(\mathcal D) = \frac{1}{N}\sum_{s_i,s_j\sim \mathcal D}\left|\|\phi_\eta(s_i) - \phi_\eta(s_j)\|_1 - d_\text{a-bisim}(s_i, s_j, \psi, \phi)\right|.
\label{bisim_loss}
\end{equation}
In practice the parameters of $\phi$ used to calculate $d_{\text{a-bisim}}$ are trailed behind $\phi_\eta$ with the exponential moving average: $\phi = \tau \phi_\eta + (1-\tau) \phi$.  



\begin{algorithm}
\textbf{Algorithm 1: Action-bisimulation Encoder Learning}
\begin{algorithmic}
  \STATE {\bfseries Input:} Dataset without reward $(s,a,s') \sim \mathcal D$, initial encoder $\phi^{\bar \theta}(s)$
  \STATE {\bfseries Single-step Training} Train $\psi$ with $\mathcal D$ and Equation~\ref{RegularizedInverseDynamics}.
  \REPEAT
      \STATE {\bfseries Forward Model Update}: Update the forward model $f_ \upsilon (\cdot)$ according to the current multi-step encoder $\phi$ using Equation~\ref{forward_model_loss}.
      \STATE {\bfseries Multi-step Update}: Sample $s_i, s_j\sim \mathcal D$ pairs and minimize the loss as defined by the metric in Equation~\ref{actionbisimulationlearn} and loss (Equation ~\ref{bisim_loss}) to update the encoder parameters $\theta$.
      \STATE {\bfseries Momentum Update}: Update the parameters: $\bar \theta = \tau \theta + (1-\tau)\bar \theta$
  \UNTIL{$\bar \theta$ converge}
  \label{algo}
\end{algorithmic}
\end{algorithm}
\section{Experiments}
\vspace{-0.1cm}
\label{experiments}
In this section, we aim to answer the following questions: \textbf{1)} Does pre-training with the action-bisimulation 
objective learn representations useful for arbitrary downstream tasks? \textbf{2)} How does this pretraining compare with existing methods, especially single-step action controllability? \textbf{(3)} Are the learned representations robust to
background distractors? \textbf{(4)} How well does the action-bisimulation procedure capture multi-step relationships between state elements? 

\begin{figure}
\centering     

\subfigure[Nav2D]{\includegraphics[width=0.22\linewidth]{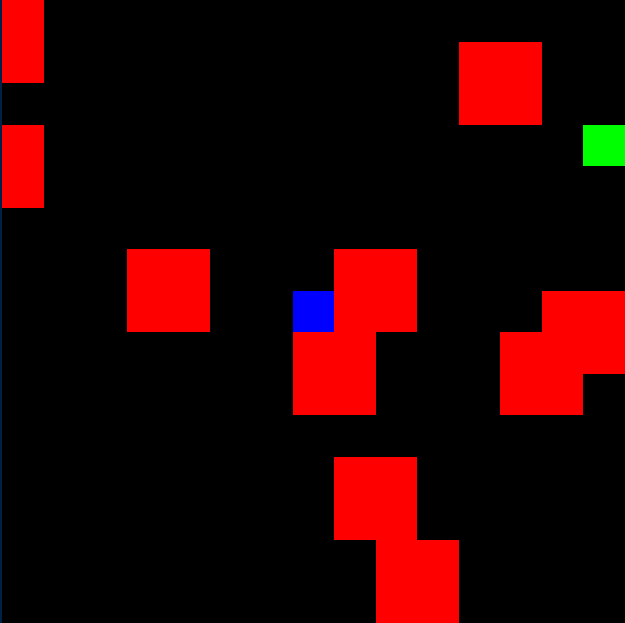}}
\hfill
\subfigure[Pointmaze]{\includegraphics[width=0.22\linewidth]{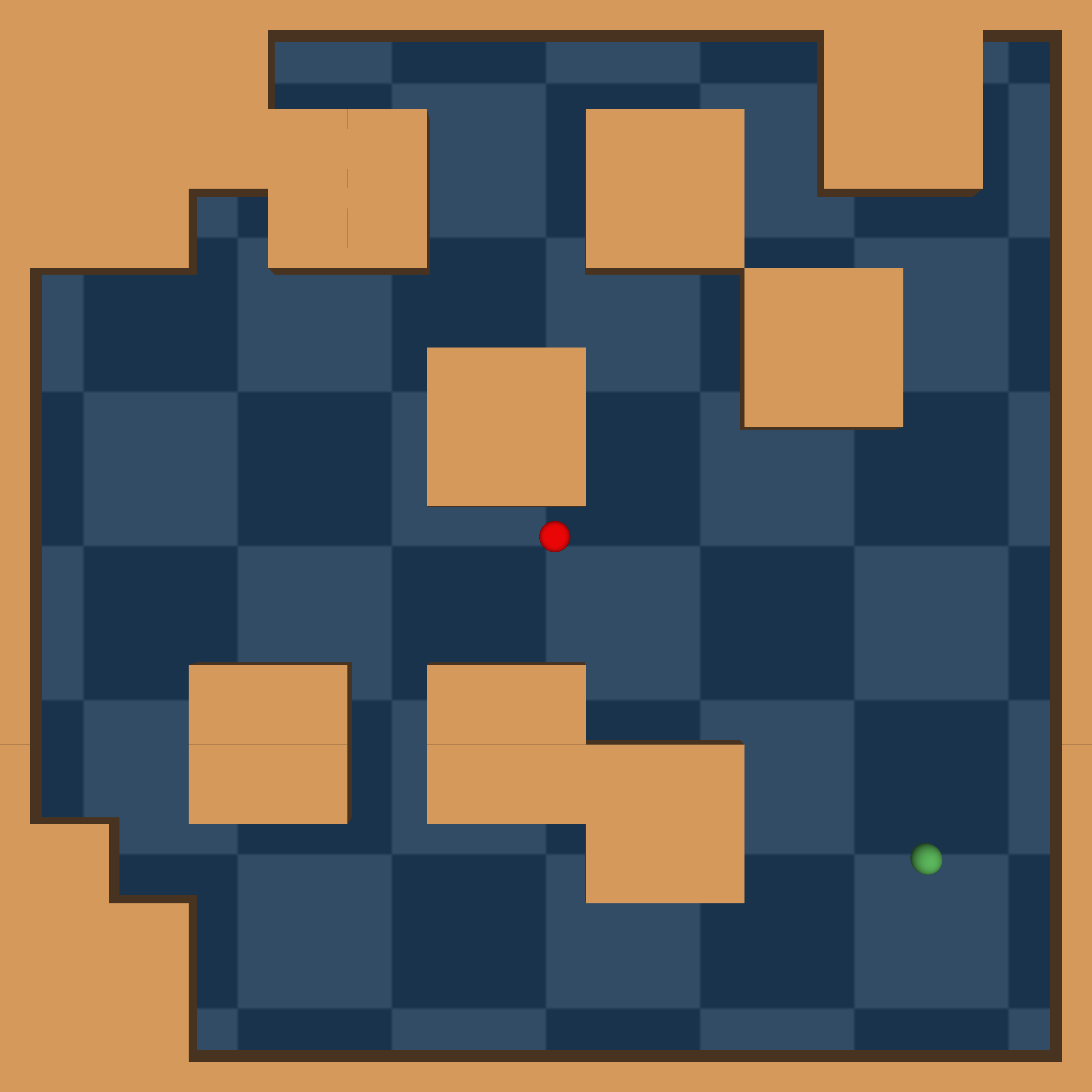}}
\hfill
\subfigure[Distractor Pointmaze]{\includegraphics[width=0.22\linewidth]{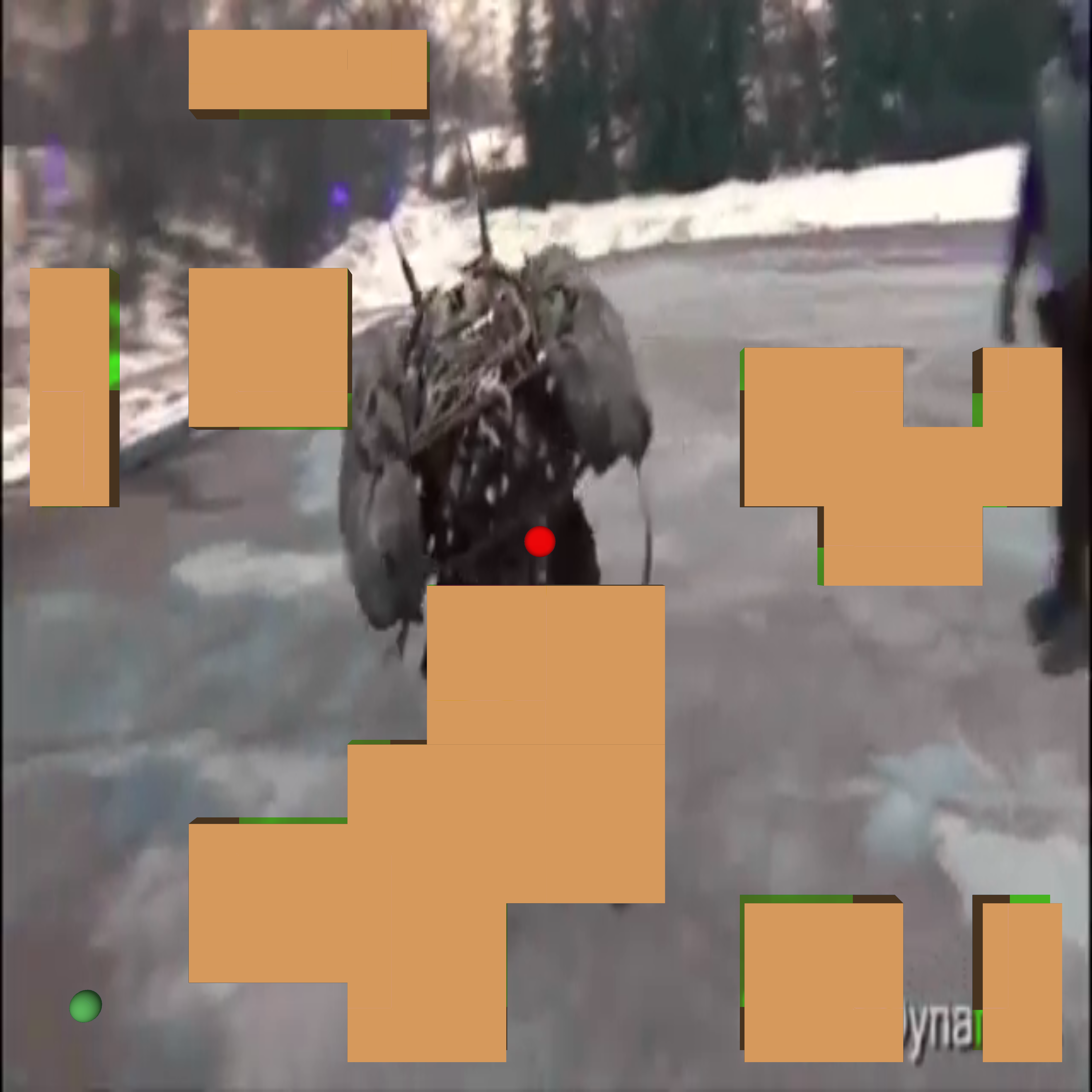}}
\hfill
\subfigure[Habitat environments]{\includegraphics[width=0.22\linewidth]{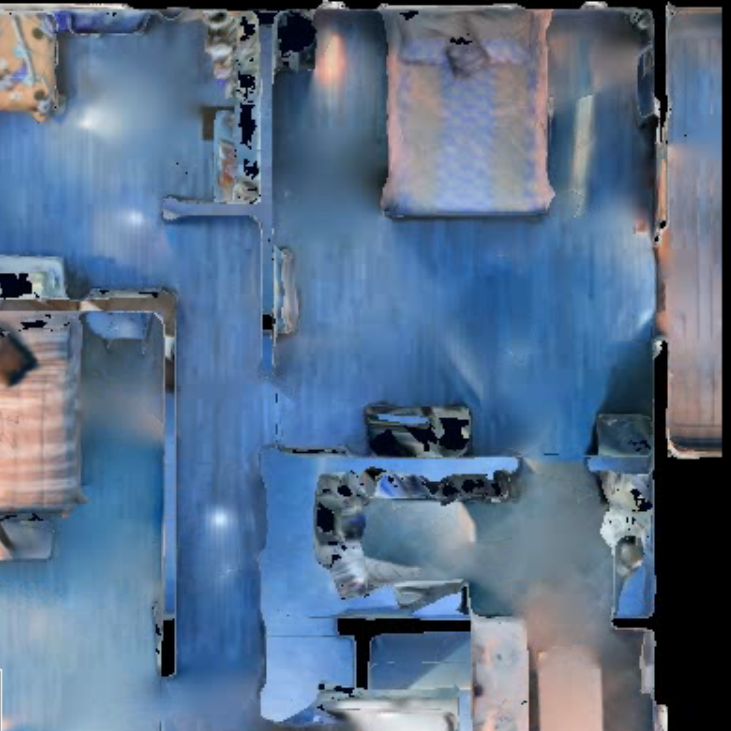}}
\vspace{-0.3cm}
\caption{ Visual representation of the RL environments.}
\vskip -0.5cm
\label{fig:environment_figure}
\end{figure}

\begin{figure}
\centering     
\subfigure[Nav2D]{\includegraphics[width=0.24\linewidth]{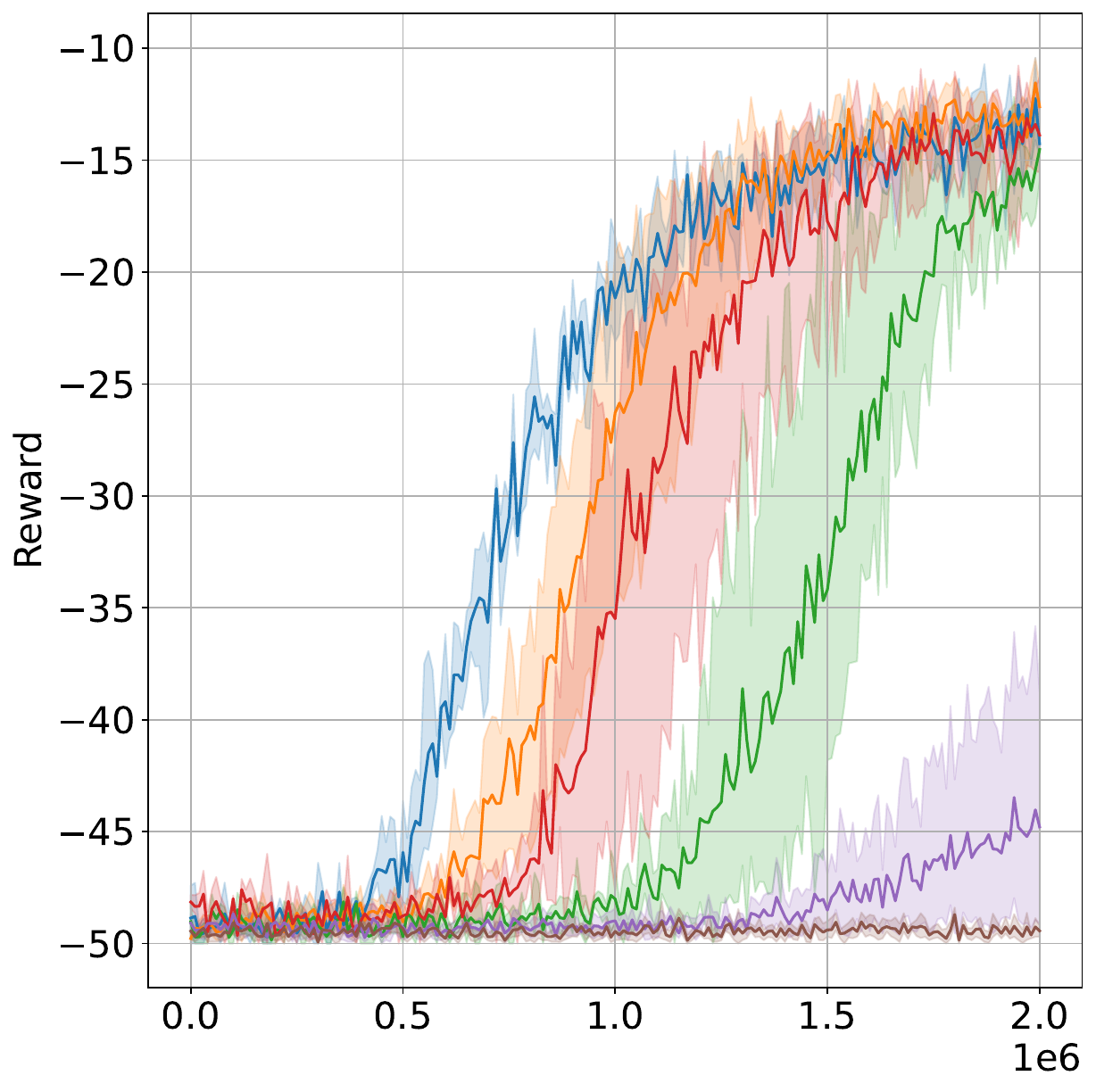}}
\subfigure[Pointmaze]{\includegraphics[width=0.24\linewidth]{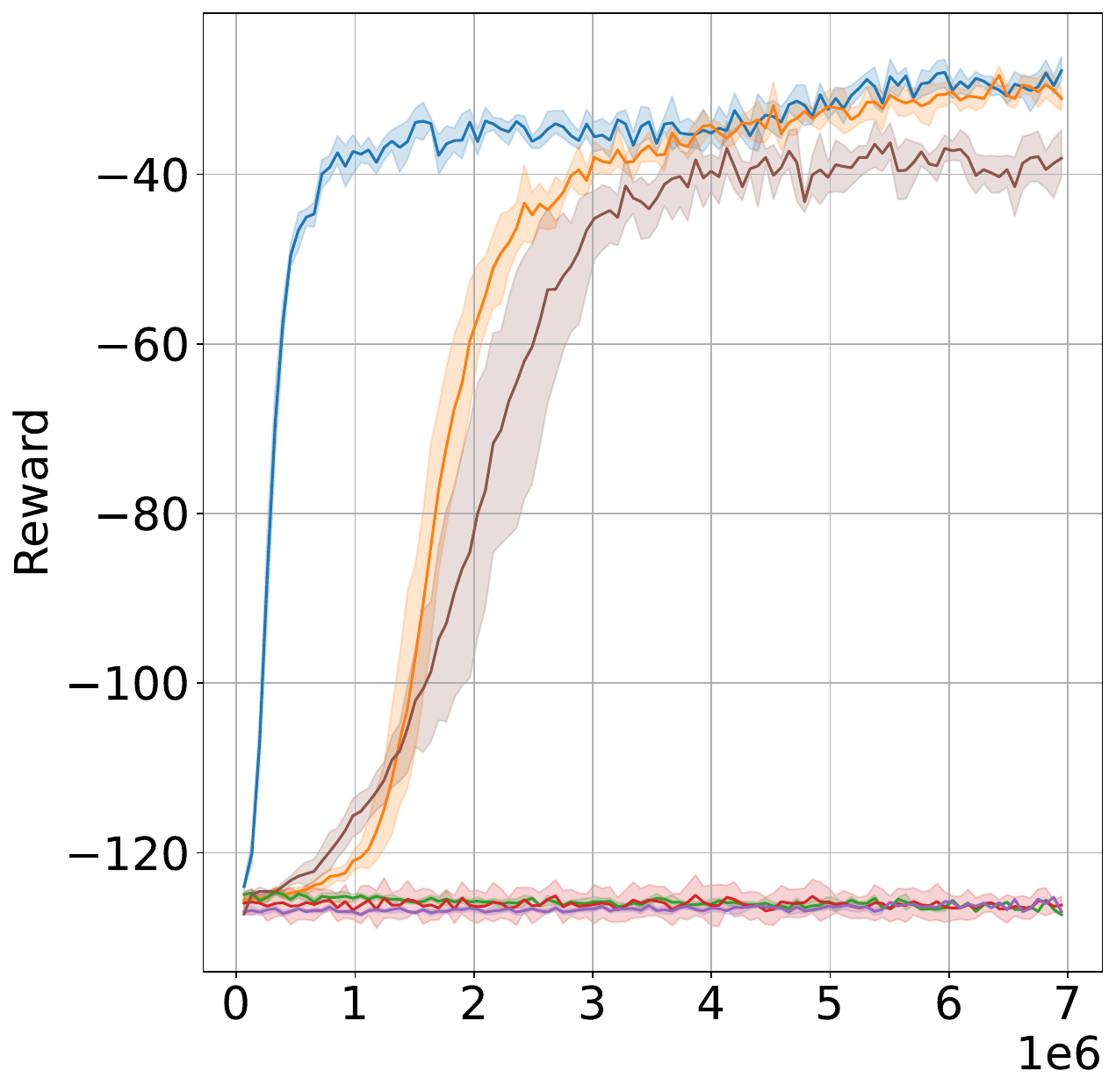}}
\subfigure[Distractor Pointmaze]{\includegraphics[width=0.24\linewidth]{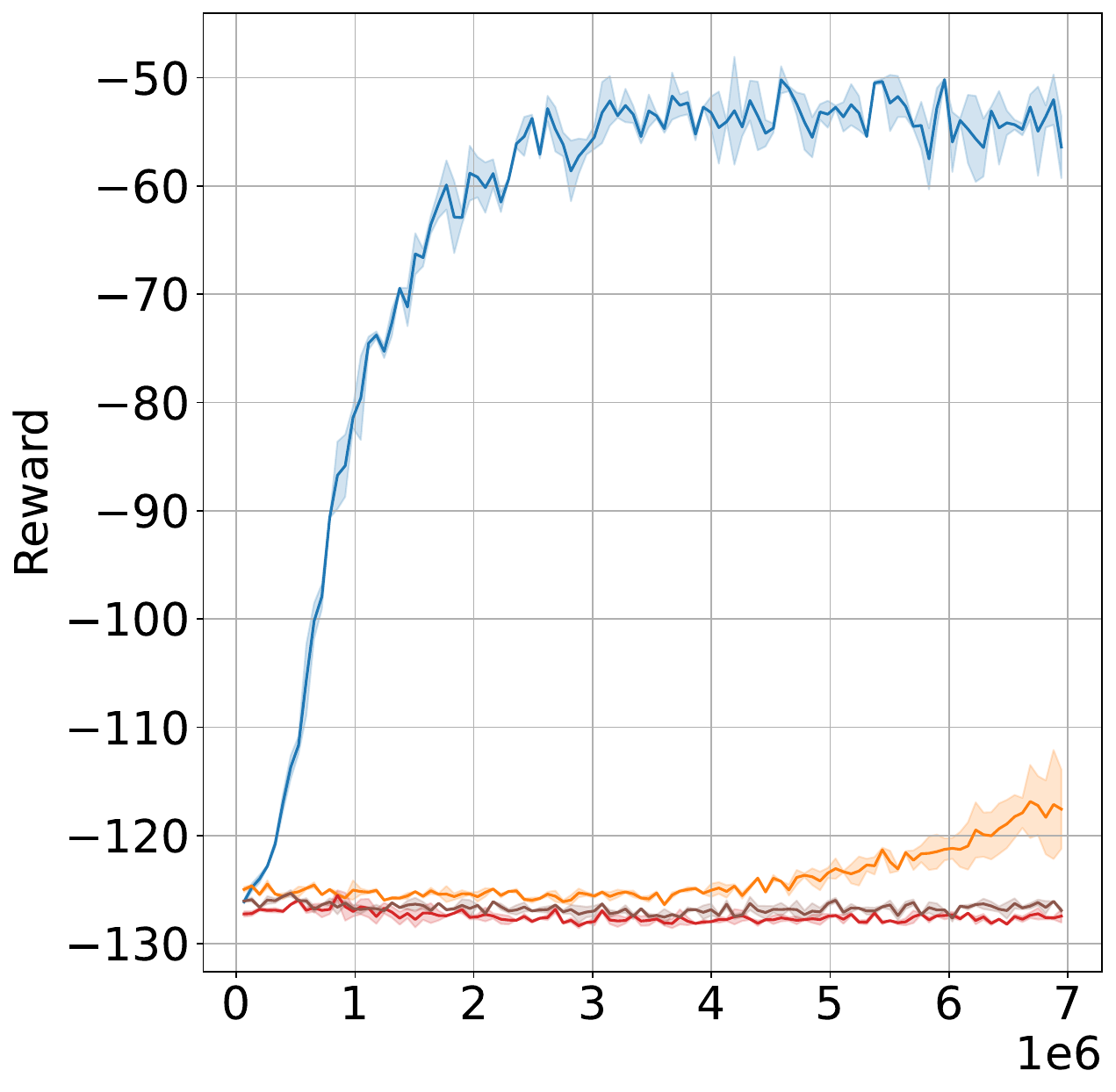}}
\vspace{-1em}
\subfigure[Habitat]{\includegraphics[width=0.24\linewidth]{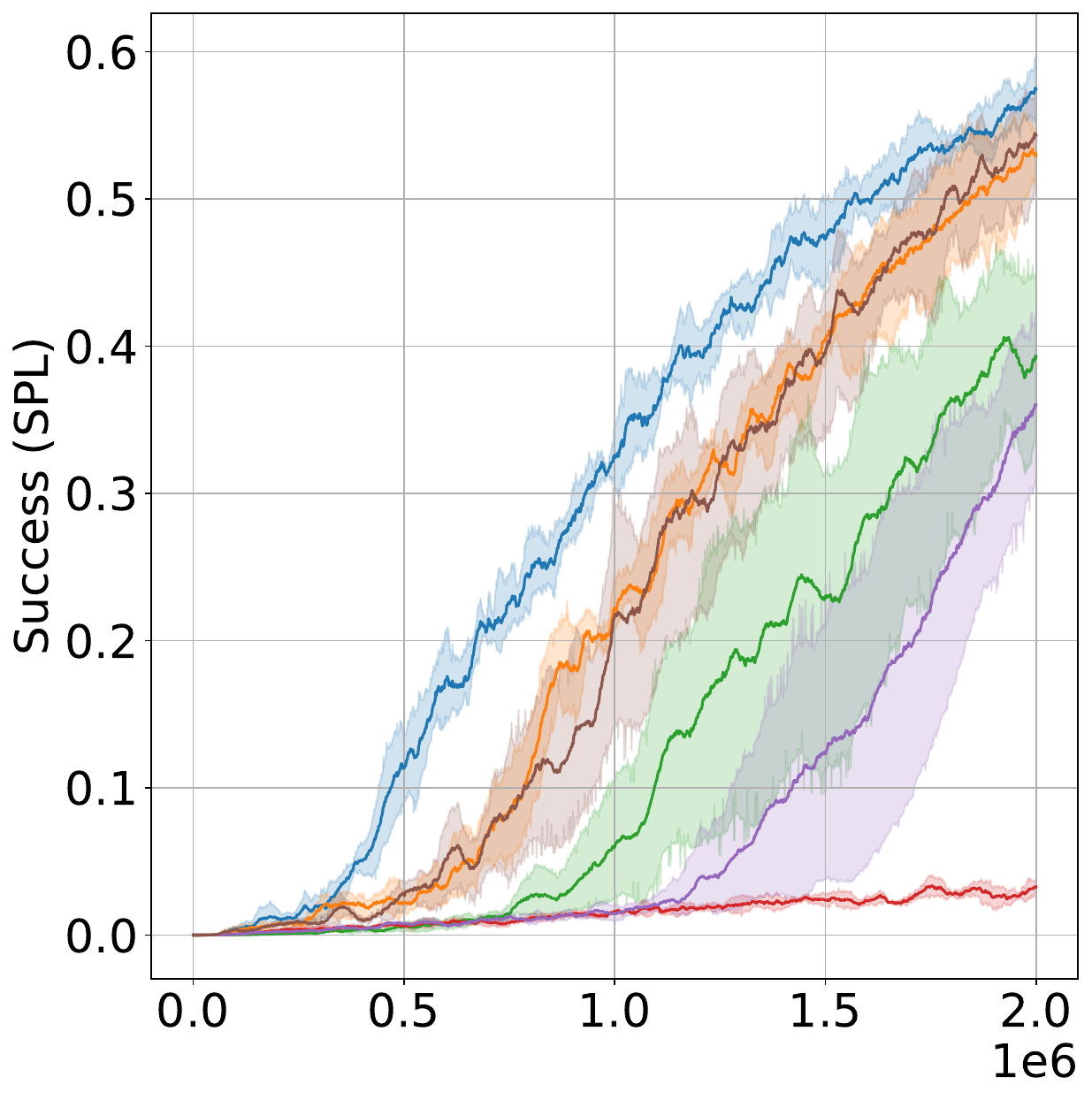}}
\vspace{-1em}
\subfigure{{\includegraphics[width=0.9\linewidth]{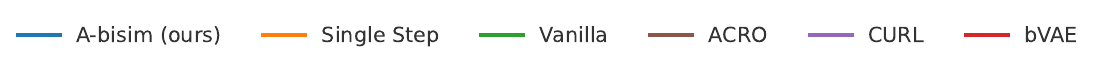}}}
\caption*{}
\vspace{-3em}
\caption{ 
Training \textbf{Performance Curves} for downstream RL learning in (a) \textbf{2D navigation}, (b) \textbf{Pointmaze navigation}, (c) \textbf{Distractor Pointmaze} and (d) \textbf{Habitat}. Mean and 95\% confidence interval are plotted over 5 trials with different random seeds for each domain.} 
\vskip -0.5cm
\label{fig:downstream_rl}
\end{figure}

We evaluate experiments in three domains illustrated in Figure~\ref{fig:downstream_rl}. \textbf{Nav2D} is a 15x15 grid environment where the agent navigates using cardinal directions to the center of the grid, avoiding randomly generated 2x2 obstacles. \textbf{Pointmaze}~\citep{fu2020d4rl} is a 2D Mujoco control environment where the agent takes actions to reach a goal location while being impeded by obstacles. We also investigate \textbf{Distractor Pointmaze} where the background in Pointmaze has been replaced with photorealistic distractions in the form of video clips. Finally, \textbf{Habitat}~\citep{savva2019habitat} is a complex 3D environment where the agent must navigate through scans of human environments to reach a goal location. Additional environment details are in Appendix~\ref{environment_appendix} (number of obstacles, goal/grid size, randomization, etc.) and all other relevant hyperparameters are in Appendix~\ref{hyperparameter_appendix}.

\subsection{Baselines}
We compare the performance of our method against representation learning pretraining methods used in prior RL works that utilize control-, contrastive- and reconstruction-based objectives.

\textbf{Single-Step Inverse} (SSI): This baseline uses the single-step objective learned using Equation~\ref{RegularizedInverseDynamics} with $k=1$ to learn a state representation. This demonstrates whether simply learning a myopic action-centric inverse dynamics representation is sufficient for good performance. In general, this representation performs surprisingly well.


\textbf{Agent Centric Representations for Offline RL} (ACRO)~\citep{lambguaranteed}: This method is equivalent to SSI with $k\neq 1$. When $k>1$, this means that the model must learn to identify the first action taken from a pair of states. While this allows the model to capture longer-term relationships, it also limits how effective it can be when trained with random actions. 

$\beta$-\textbf{Variational Autoencoder} (bVAE)~\citep{higgins2016beta}: This method evaluates a classic compressed state reconstruction method for representation learning. While popularized with video, it has been applied to RL with marginal success. In general, reconstruction can struggle to pick up fine-grained changes such as the movement of the agent.

\textbf{Contrastive Unsupervised Representations for Reinforcement Learning} (CURL)~\citep{laskin2020curl}: This method uses data augmentation with a contrastive objective to learn a representation. In this work, we used random noise augmentations because of the importance of identifying small features (the location of the agent).

\textbf{Vanilla RL}~\citep{schulman2017proximal,mnih2013playing}: Trains either Deep Q-networks (DQN) in Gridworld, or Proximal Policy Optimization (PPO) in the remaining domains, from scratch.







\subsection{Downstream Learning}
To evaluate downstream learning we first gather an offline dataset of random action state transitions, with sizes recorded in Table~\ref{DatasetTable}. State encoder $\phi(s)$ is trained with Equation~\ref{bisim_loss}, which is used to initialize the policy $\pi_\theta(\cdot|\phi(s))$. This fine-tuning strategy proved to be the best performing empirically, though future work could investigate freezing the encoder, the technique used in~\cite{lambguaranteed}, as we discuss in Appendix~\ref{alternative_training}. \cref{fig:downstream_rl} illustrates the comparison of the action-bisimulation encoder to baseline encodings. 

As we can see in Figure \ref{fig:downstream_rl}, learning with the action-bisimulation representation outperforms other methods in terms of sample efficiency, even k-step controllability (ACRO), by a substantial margin. This provides evidence for \textbf{hypotheses 1 and 2}, that action-bisimulation learns useful representations which compare well with other methods. That reconstruction and data augmentation-based methods bVAE and CURL perform poorly is not unexpected: in this domain, the agent is often small, so these methods achieve low reconstruction loss even when they omit the most important element: the agent position. On the other hand, SSI captures the agent position but is highly myopic, limiting transfer to downstream tasks. We hypothesize ACRO struggles because it relies on predicting action from two states separated by $k$ timesteps, which is ill-posed, especially when using a dataset of random actions. Additional details on baselines can be found in Appendix~\ref{baseline_details}. 


\begin{figure}
\centering     

\subfigure[Action-bisimulation]{\includegraphics[width=0.24\linewidth]{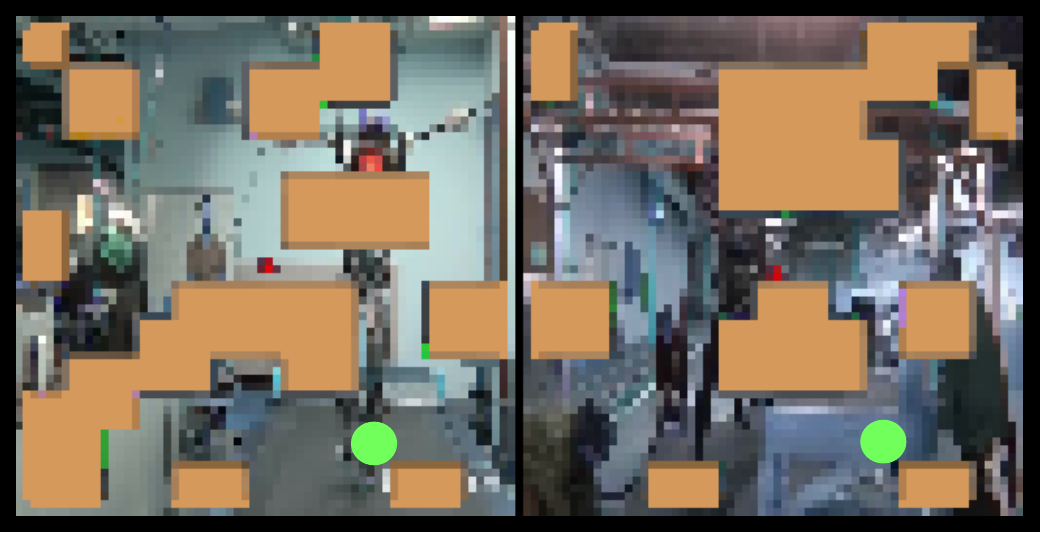}}
\subfigure[Single-Step]{\includegraphics[width=0.24\linewidth]{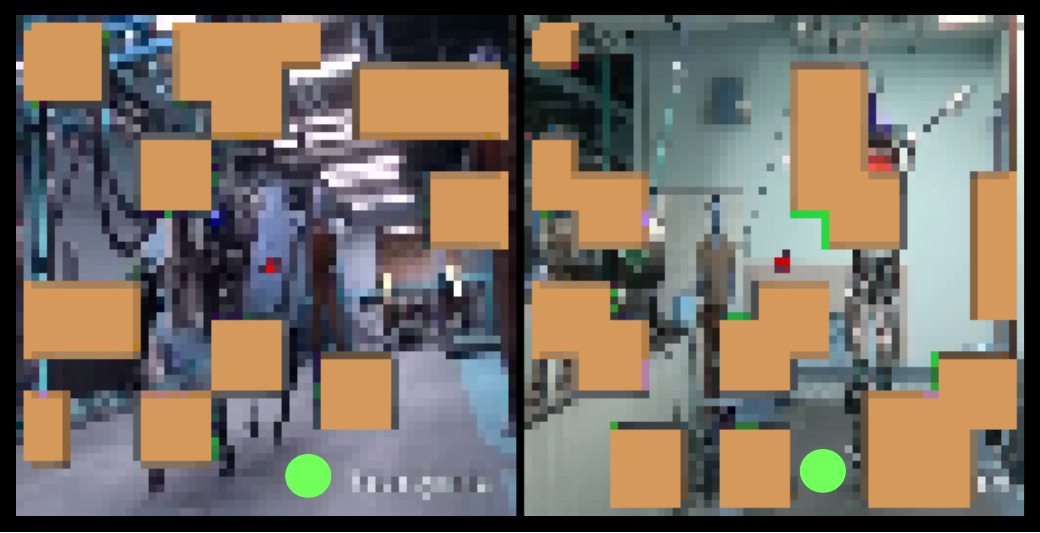}}
\subfigure[ACRO]{\includegraphics[width=0.24\linewidth]{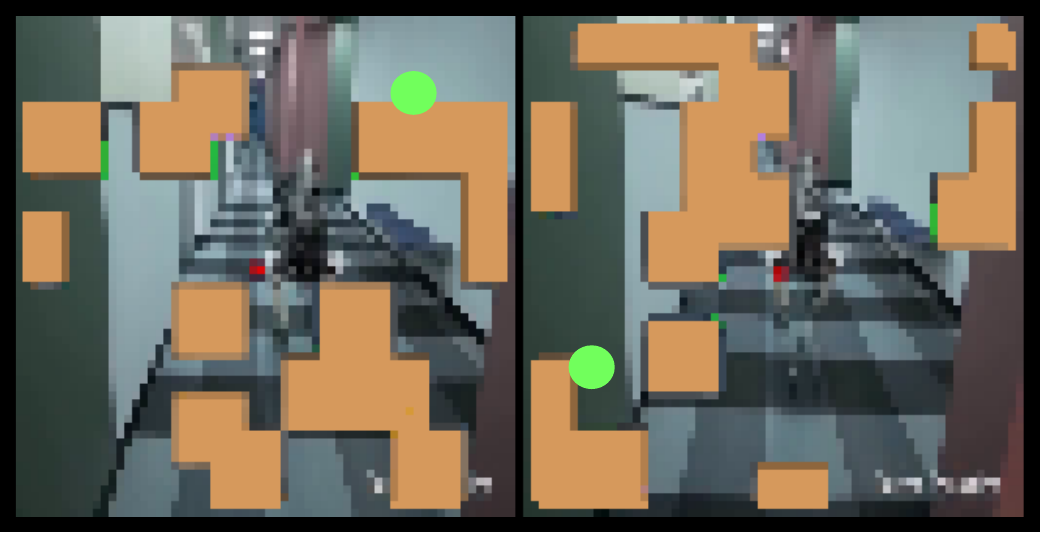}}
\subfigure[$\beta$-VAE]{\includegraphics[width=0.24\linewidth]{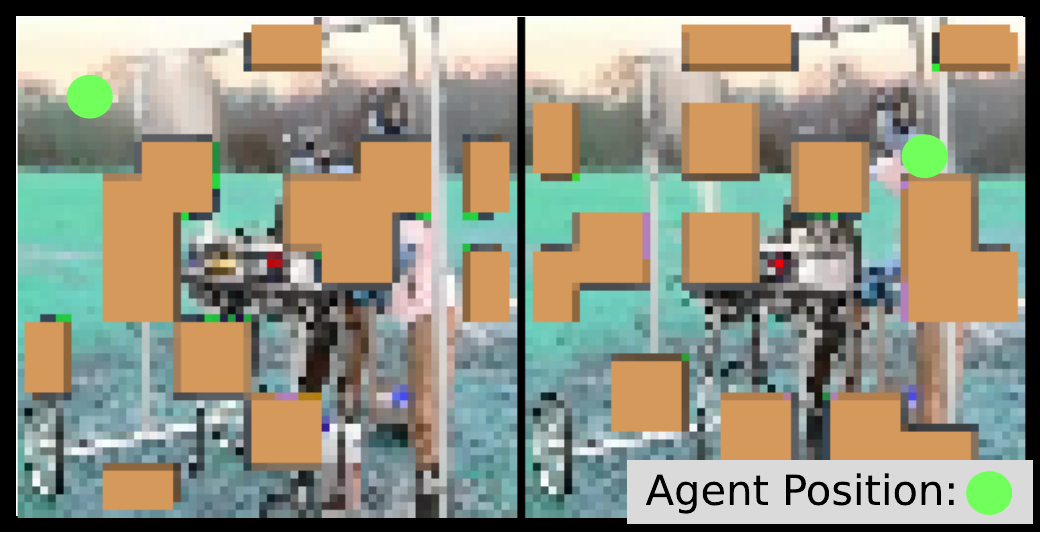}}
\caption{ 
\textbf{States close together in embedding space} drawn from the Distractor Pointmaze domain. Notice that action-bisimulation captures both the agent location and the local region of obstacles, while other methods are distracted by the background (ACRO, bVAE) or only capture one-step relations (single step). The agent is exaggerated in these images so it is easier to locate---in reality, it is quite hard to detect because of the distractors.}
\vskip -0.5cm
\label{paired_images}
\end{figure}



\subsection{Background Distractors}
In this section we evaluate \textbf{hypothesis 3}: whether the action-bisimulation encoding is robust to distractors. We assess this through a modified Pointmass environment with a photorealistic visual background. The foreground, that is the agent, goal position, and obstacles, remain the same. We visualize the distractor environments in Figure~\ref{paired_images}, where the agent has been exaggerated.

Figure~\ref{fig:downstream_rl}c shows that adding background distractors dramatically widens the gap between action-bisimulation and other methods. These backgrounds make vanilla RL, reconstruction and data-augmentation-based methods struggle wildly since these methods have no built-in robustness. They also have a significant effect, even on the fixed-step models, ACRO, and SSI. For single-step models, we hypothesize this is because pretraining causes the agent to mostly ignore obstacles since they have a limited myopic effect. For ACRO, the correlated background images appear to confuse the k-step prediction. For action-bisimulation, by contrast, there is only a marginal difference.

We also illustrate how a few representative methods map together states in Figure~\ref{paired_images}. In these plots, two nearby states are sampled and visualized. As we can see, action-bisimulation and single-step encodings encode the agent position, but action-bisimulation also maps regions of similar local obstacles together. Beta-VAE (bVAE) encodings are trained with reconstruction; the encodings largely ignore the agent in favor of matching similar backgrounds. Interestingly, ACRO also maps similar backgrounds together. We think this is because of the correlation between subsequent frames in the video, though this is worth further investigation.

\subsection{Captured Representations}
To investigate \textbf{hypothesis 4}: how well the action-bisimulation encodings capture multi-step relationships, we provide qualitative visualizations comparing the multi-step and single-step encodings.
\begin{wrapfigure}{r}{0.55\textwidth}
\vspace{-0.6cm}
\centering     
\subfigure[]{\includegraphics[width=0.4\linewidth]{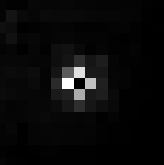}}
\subfigure[]{\includegraphics[width=0.4\linewidth]{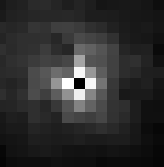}}
\vspace{-0.3cm}
\caption{ \textbf{Perturbation map of single step vs action-bisimulation} shows encoder distance in 2D Navigation when obstacles are toggled at all locations around the agent (located at the center). Brightness at a pixel indicates the size of the change of representation. \textbf{Left: The Single Step} encoder myopically captures only directly adjacent obstacles. \textbf{Right: The Multi Step} encoder captures more distant obstacles.
}
\label{fig:perturbation_maps}
\vspace{-0.6cm}
\end{wrapfigure}
Figures~\ref{fig:perturbation_maps} is a \textbf{perturbation map}, which visualizes how much the representation changes when a single obstacle is placed at a particular location, compared with the base representation. Figure~\ref{fig:perturbation_maps} illustrates the contrast between the myopia of the single-step encoder compared with the range of the multi-step encoder. 

In Appendix~\ref{qualitative_results}, we provide several additional qualitative results demonstrating how the action-bisimulation representation captures multi-step relations, including perturbation plots of how the sensitivity changes with $c$, the tradeoff parameter, and the representation difference from near-vs-far perturbations. Furthermore, sensitivity to perturbations is environment-dependent: if the environment has a fixed structure such as a corridor or maze, unreachable obstacle perturbations will be mapped close together in the action-bisimulation space.



\section{Conclusion}

Controllability-capturing encodings for reinforcement learning are a promising direction for representation pretraining since they can be learned without reward but are still able to filter out uncontrollable distractors.  However, existing methods either only capture short-term controllability or are dependent on demonstration data, which has implicit task bias. We introduce the action-bisimulation encoding, which builds off of myopic representations by enforcing recursive invariance to learn a supervision-free multi-step controllability representation. The empirical results in this work demonstrate how these encodings can be used to improve the sample efficiency, especially in domains with significant background distractors. The primary limitation of this method is the inverse dynamics single-step model, which might not capture \textit{all} controllable features, just a subset. This can result in the representation being agnostic to important task elements. A more in-depth discussion of limitations is included in Appendix~\ref{sec:limitations}. Altogether, action-bisimulation is a novel invariance relation for capturing controllability from offline data that removes expert performance requirements and smoothly handles long-horizon controllability.

\section{Acknowledgements}
This work has taken place in part in the Safe, Correct, and Aligned Learning and Robotics Lab (SCALAR) at The University of Massachusetts Amherst. SCALAR research is supported in part by the NSF (IIS-2323384), AFOSR (FA9550-20-1-0077), and the Center for AI Safety (CAIS). The work was supported by the National Defense Science \& Engineering Graduate (NDSEG) Fellowship sponsored by the Air Force Office of Science and Research (AFOSR). Special thanks to collaborators Stephen Guigere, Harshit Sikchi, Alex Levine, Siddhant Agarwal, Rudolf Lioutikov, Yuchen Cui, Akanksha Saran, Wonjoon Goo, Daniel Brown, Prasoon Goyal, Christina Yuan, and Ajinkya Jain for their fruitful conversations and timely help.
\bibliography{refs}
\bibliographystyle{rlc}


\appendix
\section{Convergence and Causal Properties}
\label{theory}
In this section, we extend some of the convergence properties that apply to reward-based bisimulation metrics to the action-based bisimulation metrics. Then, we prove that an optimized representation is agnostic to causally irrelevant components: elements that do not affect control and cannot be affected by control.

\subsection{Fixed point convergence}
First, we demonstrate that our action bisimulation metric converges to a fixed point. This proof follows a similar pattern to that found in~\cite{agarwal2021contrastive}.
\begin{theorem}
Let $\mathcal M$ be the space of bounded pseudometrics on $\mathcal S, \mathcal A$. Define operator $\mathcal F: \mathcal{M} $ based on the action-bisim distance metric in Theorem~\ref{actionbisimulation}: \begin{align}\mathcal F(d)(s_i, s_j) &= d_\text{ss}(s_i, s_j) +c \cdot E_{a\sim U(\mathcal A)}[W_1(d)(P(\cdot|s_i,a), P(\cdot|s_j,a))].\nonumber \end{align}
Then $\mathcal F$ is a contraction mapping and has a unique fixed point for a bounded dist.
\label{FixedPointTheory}
\end{theorem}

\textbf{Proof}: See Appendix~\ref{FixedPointProof}. $\blacksquare$

\subsection{Agnostic to Behavior Irrelevant Components}
Just because there is an optimal fixed point does not imply that this optimal fixed point is useful. Even using a trivial single-step embedding $\psi$ which maps all states to zero will still satisfy the convergence. However, if we assume that $\psi(s)$, the single-step representation, captures only action-relevant information between $S$ and $S'$, the myopic state information, then we can show that the learned representation captures a subset of the control relevant state features only. 

First, we assume a uniform behavior policy:
\begin{assumption}
\label{uniformpolicy}
The distribution of $\pi(a|s)$ is uniform (uniform distribution denoted $U(\mathcal A)$), and therefore not conditioned on $S$:
$$P(a|S)= \frac{1}{|\mathcal A|} \quad \forall a$$. 
\end{assumption}
This is because otherwise, the behavior policy could introduce relationships between states and actions that are not present as a result of control. Now we turn to the properties of the single-step encoder. Using the abuse of notation where $\psi(S)$ is the random variable representing state, we make the following assumption about the single-step model:
\begin{assumption}
\label{restrictedactionencoder}
$\psi: \mathcal S \rightarrow \mathcal Z_\text{ss}$ captures a minimum sufficient representation between $S, S'$ and $A$:
\begin{align}
\psi \coloneqq \argmin_\psi &\quad I\left(S;\psi(S)\right) \nonumber\\
& \text{s. t.} \quad d_{KL}\left(P\left(A|[\psi(S), \psi(S')]\right)\| P\left(A|[S, S']\right)\right) = 0,
\label{ActionInfoBottleneck}
\end{align}
\end{assumption}

where $d_{KL}(\cdot\|\cdot)$ is the KL divergence between two distributions. Then this question denotes that $\psi(s)$ captures as little information about the current state as possible (the first term), the conditional distribution over $A$ from $[\psi(S), \psi(S')]$ is the same as that using $[S, S']$. Notice that the terms in this assumption are approximated in single step encoder training (Equation~\ref{RegularizedInverseDynamics}). The inverse dynamics prediction approximates the KL constraint, and the encoding regularization ensures minimal remaining information.

Before using this assumption, we first define what kind of information our representation should be agnostic to. Suppose that there is a partitioning of the state features (analogous to causal feature sets in~\citep{zhang2020invariant}) where one set is controllable $S^c$, and any feature not part of that set is $S^u$. The sets can be imagined as sets of causal variables, where the concatenation of these sets produces the complete state space $S$. These sets can be defined as follows: 
\begin{definition}
\label{controllablepartition}
State $S$ can be decomposed into controllable feature set $S^c$ and uncontrollable feature set $S^u$ that completely describe $S$ (bidirectional entropy is $1$). These partitions have the property that the transition dynamics of $S^c$ are independent of the transition dynamics of $S^u$, and the transition dynamics of $S^u$ are independent of $S^c$ and $A$:
\begin{align}
P(S^{u'}|S,A) = P(S^{u'}|S^u)\nonumber\\
P(S^{c'}|S,A) = P(S^{c'}|S^c,A)\nonumber \\
H(S|S^c, S^u) = H(S^c, S^u|S) = 1.
\label{controllablepartitioneq}
\end{align}
\end{definition}

The encoder will compress action-irrelevant components (elements of $S^u$), which are components with no undirected path in the causal graph connected to actions. By compression, we mean that states that vary only according to these elements will share the same encoding.

\begin{theorem}
\textbf{Action-Bisimulation Control Relevance:} Suppose that $\phi: \mathcal S \rightarrow \mathcal Z$ maps observations to a latent action bisimulation representation where $\|\phi(s_i) - \phi(s_j)\|_1 = d_\text{a-bisim}(s_i, s_j, \psi,\phi)$ using a $\psi$ described in Definition~\ref{restrictedactionencoder}. $Z$, the distribution of encodings has no information about action-irrelevant components: $I(Z; S^u) = 0$.
\label{regularized_theory}
\end{theorem}
\textbf{Proof}: See Appendix~\ref{regularizedproof}. $\blacksquare$

\newtheorem*{repro}{\textbf{Theorem (reproduced)}}
\newtheorem*{reprolem}{\textbf{Lemma (reproduced)}}

\section{Proofs}
\subsection{Fixed point proof}
\label{FixedPointProof}
\begin{repro}
Let $\mathcal M$ be the space of bounded pseudometrics on $\mathcal S, \mathcal A$. Define operator $\mathcal F: \mathcal{M} $ based on the action-bisim distance metric in Theorem~\ref{actionbisimulation}: \begin{align}\mathcal F(d)(s_i, s_j) &= d_\text{ss}(s_i, s_j) +c \cdot E_{a\sim U(\mathcal A)}[W_1(d)(P(\cdot|s_i,a), P(\cdot|s_j,a))].\nonumber \end{align}
then $\mathcal F$ is a contraction mapping and has a unique fixed point for a bounded dist.
\end{repro}

\textbf{Proof}:\\
First, we utilize a lemma that is proved in ~\citet{agarwal2021contrastive}, which allows us to apply a powerful inequality to the bisimulation-esque pseudometric defined in Equation~\ref{actionbisimulation} in Appendix~\ref{theory}.

\begin{lemma} 
\label{pseduometricineq}
Inequality for two pseudometrics $d,d'$ and probability distributions $P_X, P_Y$:
\begin{equation}
W_1(d)(P_X, P_Y) \leq \|d - d'\| + W_1(d')(P_X,P_Y ).
\label{PSMineq}
\end{equation}
\end{lemma}
See \cref{pseduometricineq} proof in~\cite{agarwal2021contrastive}.

Then, use Banach fixed point theorem:
\begin{align}
\mathcal F(d)(x, y) &- \mathcal F(d')(x, y) = \nonumber \\ 
&= c \cdot E_{a \sim U(\mathcal A)}[W_1(d)(P(\cdot|s_i,a), P(\cdot|s_j,a))] - E_{b \sim U(\mathcal A)}[W_1(d')(P(\cdot|s_i,b), P(\cdot|s_j,b))] \nonumber \\
&= c \cdot E_{a \sim U(\mathcal A)}[W_1(d)(P(\cdot|s_i,a), P(\cdot|s_j,a))] - W_1(d')(P(\cdot|s_i,b), P(\cdot|s_j,b))] \nonumber \\
& \mathop{\leq  c \cdot E_{a \sim U(\mathcal A)}[\|d - d'\| + W_1(d')(P(\cdot|s_i,a), P(\cdot|s_j,a))] - W_1(d')(P(\cdot|s_i,b), P(\cdot|s_j,b))]}_{\text{Applying Lemma~\ref{PSMineq}}}. \nonumber \\ 
&= c \cdot E_{a \sim U(\mathcal A)}[\|d - d'\|] \nonumber \\
&= c \cdot \|d - d'\| \nonumber
\end{align}
Since $\mathcal F(d)(x, y) - \mathcal F(d')(x, y) \leq c \cdot \|d - d'\|$, $\mathcal F$ is a contractive mapping for $c < 1$ and has unique fixed point $d^{*}.$ $  \blacksquare$

\subsection{Causal Parititon proof}
\label{featuresetproof}
\begin{assumption}
$\psi$ captures the information bottleneck representation between $S^t, S^{t+k}$ and $A_k$:
\begin{equation}
\argmin{\psi} I(S^t,S^{t+k};\psi(S), \psi(S^{t+k})) - \beta I(\psi(S), \psi(S^{t+k});A_k)
\label{ActionInfoBottleneck}
\end{equation}
\end{assumption}
Then, the following theorem holds:

\begin{theorem}
\textbf{Action Bisimulation Partitions:} If we partition observations using the action bisimulation metric where the single-step representation optimizes Equation~\ref{ActionInfoBottleneck}, then the action bisimulation partitions correspond to a subset of the causal feature set for current and future actions.
\end{theorem}

\textbf{Proof}: \\
Suppose $u$ is a feature along which action bisimulation partitions, but is not part of the causal feature set for current and future actions. 

First, consider the case of current actions: then by definition, this will increase $I(S^t,S^{t+k};\psi(S), \psi(S^{t+k}))$, because it will be a component of state encoded by the embedding. However, since it is not part of the causal feature set, it will not increase $\beta I(\psi(S), \psi(S^{t+k});A_k)$. Thus, it will not satisfy the optimal embedding specified in Equation~\ref{ActionInfoBottleneck} for the single step embedding, which will increase the base-case loss in Equation~\ref{bisim_loss}, $\|\psi(s_i) - \psi(s_j)\|$. 

Second, consider that in the case where $u$ encodes information about future actions, suppose at time horizon $k$. This will increase the loss in the second term for Equation~\ref{bisim_loss}. This can be seen by unrolling the distance across $k$ steps, where the l1 loss is used in the Wasserstein distance. 

Thus, $u$ cannot exist while also being an optimal solution, meaning it could not be a feature along which the action bisimulation partitions. $\blacksquare$

This connection allows us to make statements about what information the encoder compresses. The encoder will compress action-irrelevant components, which are components with no undirected path in the causal graph connected to actions so that these states are encoded together.

\subsection{Action-Bisimulation Control Relevance Proof}
\label{regularizedproof}
\begin{repro}
\textbf{Action-Bisimulation Control Relevance:} Suppose that $\phi: \mathcal S \rightarrow \mathcal Z$ maps observations to a latent action bisimulation representation where $\|\phi(s_i) - \phi(s_j)\|_1 = d_\text{a-bisim}(s_i, s_j)$. $Z$, the distribution of encodings has no information about action-irrelevant components: $I(Z; S^u) = 0$.
\end{repro}

\textbf{Proof}: \\
$s_i$ and $s_j$ are two states which only differ according to features in $S^u$. We demonstrate for any $s_i,s_j,S^u$, $\phi(s_i) = \phi(s_j)$.

\begin{lemma}
\label{zeroMISS}
$I(\psi(S);S^u) = 0$ for any $\psi(\cdot)$ that satisfies \cref{restrictedactionencoder}
\end{lemma}
\textbf{Proof of }\cref{zeroMISS}: \\
We start by demonstrating that a $\psi(S)$ with zero mutual information with $S^u$ can still satisfy $d_{KL}\left(P\left(A|[\psi(S), \psi(S')]\right)\| P\left(A|[S, S']\right)\right) = 0$, the distribution matching property of Equation~\ref{ActionInfoBottleneck} by demonstrating that the distributions have no dependence on $S^u$:
\begin{align}
P(A|S,S') &= \frac{P(S'|S,A)P(A|S)}{P(S'|S)}\nonumber\\
 &= \frac{P(S^{u'}|S,A)P(S^{c'}|S,S^{u'},A)P(A|S)}{P(S^{u'}|S)P(S^{c'}|S,S^{u'})} \nonumber\\
&= \mathop{\frac{P(S^{u'}|S^u)P(S^{c'}|S^c,A))P(A|S)}{P(S^{u'}|S^u)P(S^{c'}|S^c)}}_{\text{Applying \cref{controllablepartition}}}\nonumber\\
&= \mathop{\frac{P(S^{c'}|S^c,A))U(\mathcal A)}{P(S^{c'}|S^c)}}_{\text{Applying \cref{uniformpolicy}}}
\label{distractorremoveequation}
\end{align}
Where $U(\mathcal A)$ is the uniform distribution over actions. By removing the dependence of $P(A|S,S')$ on $S^u$, this means that $d_{KL}(P(A|S,S')\|P(A|\psi (S),\psi (S'))) = 0$ for all $\phi(S)$ where the distribution differs only according to $S^u$.

Now, consider any $\tilde \psi(S)$ where $I(\tilde \psi(S);S^u) = \alpha > 0 $. We have already shown that the $\tilde \psi(S)$ distributional dependence is unnecessary to satisfy the $KL$ constraint. Thus, any dependence on $S^u$ will increase the mutual information $I(S;\tilde \psi(s))$. This means that for any single step encoding $\tilde \psi(\cdot)$, there exists a lower cost $\psi(\cdot)$ which has no dependence on $S^u$, since any dependence on $S^u$ is unnecessary to satisfy the KL constraint. Thus any $\psi(\cdot)$ that satisfies~\cref{restrictedactionencoder} has the property $I(\psi(S);S^u) = 0 $.

\begin{lemma}
\label{zeroSSDist}
For any $s_i, s_j \in S$ which differ only according to features in $S^u$, $$\|\psi(s_i) - \psi(s_j)\| = 0$$
\end{lemma}

\textbf{Proof of }\cref{zeroSSDist}:\\ 
The consequence of \cref{zeroMISS} is that the zero mutual information indicates: 
$$\psi(s_i) = \psi(s_j) \quad \quad \forall s_i, s_j\quad \text{s. t.} \quad \text{$s_i$ and $s_j$ differ only according to features in $S^u$}$$
This follows from the definition of mutual information, where $I(X,Y) = 0$ implies that $X$ is independent of $Y$. If two variables are independent, then any change of one variable will not change the other variable. As a result, $\|\psi(s_i) - \psi(s_j)\| = 0$. $\blacksquare$

Finally, we can complete the proof by unrolling the multi-step objective for any two states $s_i, s_j$ which differ only according to $S^u$:
\begin{align}
d_\text{a-bisim}(s_i, s_j, \psi, \phi) =  (1-c)\cdot\|\psi_(s_i) - \psi_(s_j)\|_1 
 + c \cdot \mathbb{E}_{a\sim U(\mathcal A)} \left[ W_1(p(\phi(s_i), a), p(\phi(s_j), a)) \right] \nonumber \\
 = c \cdot \mathbb{E}_{a\sim U(\mathcal A)} \left[ W_1(p(\phi(s_i), a), p(\phi(s_j), a)) \right] \nonumber \\
 = c \cdot E_{a\sim \mathcal U(\mathcal A)}\left[\int_{s_i'\sim p(\phi(s_i), a), s_j'\sim p(\phi(s_j), a)} d_\text{a-bisim}(s_i',s_j') \delta s_i' \delta s_j'\right]\nonumber 
\end{align}
Notice that unrolling $d_\text{a-bisim}(s_i',s_j')$ gives $(1-c)\cdot\|\psi_(s_i') - \psi_(s_j')\|_1 
 + c \cdot \mathbb{E}_{a\sim U(\mathcal A)} \left[ W_1(p(\phi(s_i'), a), p(\phi(s_j'), a), d) \right]$. Using \cref{controllablepartition} demonstrates that $s_i'$ and $s_j'$ must also only differ according to features in $S^u$. By induction, this difference holds for all timesteps, which demonstrates that $d_\text{a-bisim}(s_i, s_j, \psi, \phi)= 0$. Since $\phi(\cdot)$ is defined as matching $d_\text{a-bisim}(s_i, s_j, \psi, \phi)$, this implies that $\phi(s_i) = \phi(s_j)\quad \forall s_i,s_j \in S$ that differ only according to $S^u$. $\blacksquare$

\section{Alternative Base Case Representations}
This section introduces the single step contrastive alternative to the encoder introduced in \cref{sec:controllability_measures}, as well as the k-step generalization, where the existing methods can be seen as $k=1$. 
\subsection{Contrastive Representations}
\label{constrativerep}
Contrastive representations approximate the lower bound of the mutual information between two signals, in this case the state transition $(s,s')$ and the action $a$. Mutual information is the degree to which knowledge about $(s,s')$ encodes information about $a$, which is defined by: $I((s,s');a) = H((s,s')) - H((s,s')|a)$, where $H$ is the Shannon entropy.
InfoNCE~\citep{oord2018representation} is a popular contrastive method for computing a lower bound of this statistic based on Noise Contrastive Estimation~\citep{gutmann2010noise}. Like inverse dyanmics, define the learned state encoder as $\psi_\theta(s): \mathcal S \rightarrow \mathcal Z_{ss}$. Define action encoder to map to the concatenated space of state encodings $[\mathcal Z_{ss}, \mathcal Z_{ss}]$, where square brackets represent concatenation: $\psi_{\eta,\mathcal A}(a): \mathcal A \rightarrow [\mathcal Z_{ss}, \mathcal Z_{ss}]$. Finally, a pairwise distance operator $d(z_1, z_2): \mathcal Z_{ss} \times \mathcal Z_{ss} \rightarrow \mathbb R$. In our experiments $d(\cdot, \cdot)$ was the l2 distance. The InfoNCE objective is as follows:
\begin{equation}
L_\text{infoNCE}(\mathcal D, \theta, \eta) = E_{(s,a^{+},s')\sim \mathcal D}\left[\frac{e^{d([\psi_\theta(s), \psi_\theta(s')], \psi_{\eta,\mathcal A}(a^{+}))}}{\sum_{\tilde a \in \{ a^{- }, a^{+}\}} e^{d([\psi_\theta(s), \psi_\theta(s')], \psi_{\eta,\mathcal A}(\tilde a))}}\right].
\label{ActioninfoNCE}
\end{equation}

$a^{+}$ denotes the positive sample, which is the actual action taken in state $s$. $a^{-}$ represents the negative samples, which are the alternative actions not taken in $s$. Optimizing the loss in Equation~\ref{ActioninfoNCE} will learn a representation encoding action-relevant components. In practice, the contrastive representations did not perform as well as the inverse dyanmics-based ones, and future work is investigating the reason for this in detail.

\subsection{K-step Base Cases for Action-Bisimulation}
\label{kstepmodels}
In this work, we primarily investigate a base encoder $\psi_\theta(\cdot)$ trained using $s,a,s'$, where $s$ and $s'$ are subsequent states. Prior work~\citep{lambguaranteed} has investigated training encoders two states $k$ steps apart and predicting the first action. While it may seem like longer-term controllability can be captured by simply increasing $k$, choosing a fixed horizon introduces a clear limitation: Determining the inverse dynamics between states when $k$ is small is well-defined but myopic, but when $k$ gets large, there may no longer be enough information between the state at $t$ and $t+k$ to provide meaningful information about the action. For this to be well defined, there must be a meaningful correlation between the current action and the state $k$ steps into the future. This correlation does not exist if the actions are random and the agent can return to states that it has been to before. As a result, in practice $k$-step controllability is limited to the offline RL setting, where some meaningful trajectories are provided to the agent~\citep{islam2022agent}. This means that in practice $k$-step methods are not fully unsupervised.

Depending on the nature of the offline data, the k-step extension can be combined with Action-bisimulation, where instead of choosing a large $c$ (i.e. $c > 0.9$), the single-step encoders can be replaced with $k$-step encoders. This has the potential to significantly increase the degree to which the action-bisimulation encoder $\phi_\eta$ can capture long-term controllability. 

Formally, instead of the tuple $(s,a,s')$, we use the tuple $(s^{(t)},a^{(t)},s^{(t+k)})$. Then, we can represent the $k$-step regularized inverse dynamics loss (adapting from \cref{RegularizedInverseDynamics}) with:
\begin{equation}
\begin{aligned}
L_{ssr}(\mathcal D, \theta, \eta) = -\sum_{(s^{(t)},a^{(t)},s^{(t+k)})\sim \mathcal D}\log f_{\eta,\text{forward}}(a^{(t)}|&\psi_\theta(s^{(t)}), \psi_\theta(s^{(t+k)})) \nonumber \\
& + \beta \left(\|\psi_\theta(s^{(t)})\|_1 + \|\psi_\theta(s^{(t+k)})\|_1\right).
\end{aligned}
\label{KRegularizedInverseDynamics}
\end{equation}

Notice that if a large $k$ is chosen, this can run into the same issues as other fixed-$k$ methods, where the distribution of actions can affect the features captured by the single step model. 

Similarly, we can replace the InfoNCE representation by replacing $a^+$ with $\mathbf a^+$, which is the actual \textit{sequence} of actions between $s^{(t)}$ and $s^{(t+k)}$, instead of just the first action. We can also replace $a^-$ with $\mathbf a^-$, which is a sequence of actions different from the actual one. This gives the $k$-step representation of \cref{ActioninfoNCE}:
\begin{equation}
L_\text{infoNCE}(\mathcal D, \theta, \eta) = E_{(ss^{(t)},\mathbf a^{+},s^{(t+k)})\sim \mathcal D}\left[\frac{e^{d([\psi_\theta(s^{(t)}), \psi_\theta(s^{(t+k)})], \psi_{\eta,\mathcal A}(\mathbf a^{+}))}}{\sum_{\tilde{\mathbf a} \in \{\textbf a^{- }, \textbf a^{+}\}} e^{d([\psi_\theta(s^{(t)}), \psi_\theta(s^{(t+k)})], \psi_{\eta,\mathcal A}(\tilde{\mathbf{a}}))}}\right].
\label{KActioninfoNCE}
\end{equation}

\subsection{Adaptive Regularization for Minimal Representation}
To train an encoder with the loss described in Eq. \ref{RegularizedInverseDynamics}, it is necessary to choose a regularizing constant $\beta$ beforehand. We found it was possible (and sometimes easier from a hyper-parameter search perspective) to adapt the $\beta$ parameter to the current performance of encoder $\psi_\theta$. We changed $\beta$ throughout training according to the accuracy of the inverse dynamics predictions, lowering the regularization constant when the accruacy was low and raising it when the accuracy was high. The intuition is that if accuracy is low, then the representation needs to be less minimal and so we need to regularize less heavily. We calculated the regularization constant $\beta_i$ where $i$ is the training iteration with:
$$\beta_i = \beta_\text{max}(1 - \exp( -4\alpha_{i-1}^2)), $$
where $\beta_\text{max}$ is the maximum regularization constant and $\alpha_{i-1}$ is the action prediction accuracy during the previous iteration. This trick did not significantly impact our results, but lessened the hyper-parameter search.

\section{Additional Qualitative Results}
\label{qualitative_results}
This section describes several other qualitative results that demonstrate the properties of the encodings learned using action-bisimulation as compared to other encoding methods. We first provide qualitative results describing how the representation is sensitive not only to the agent's location but also to the local obstacles. This distinction is valuable since encoding agent position can often be sufficient to already significantly improve downstream RL performance. To generate the plot, we randomly generate obstacles either near the agent (left) or far from the agent (right), where near and distance are described below. When the changes are near the agent, there is a large variation in representation distance. On the other hand, distant perturbations make little difference to the representation of the agent. 
\begin{figure}[H]
\centering     
\subfigure[]{\includegraphics[width=0.3\linewidth]{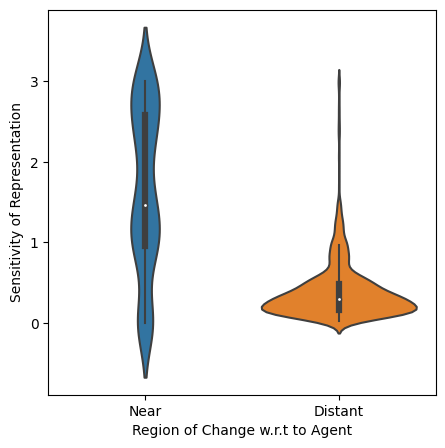}}
\subfigure[]{\includegraphics[width=0.30\linewidth]{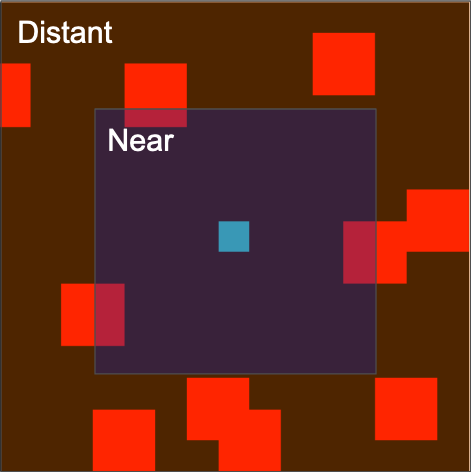}}
\caption{ \textbf{Left: Violin Plot} shows how the representation is sensitive to changes in obstacles near and distant to the agent. \textbf{Right: Sample observation} illustrates the \textit{near} and \textit{distant} regions with respect to the agent in the center.
}
\vskip -0.5cm
\label{fig:violin_sensitivity}
\end{figure}

\begin{figure}[H]
\centering     
\subfigure[]{\includegraphics[width=0.2\linewidth]{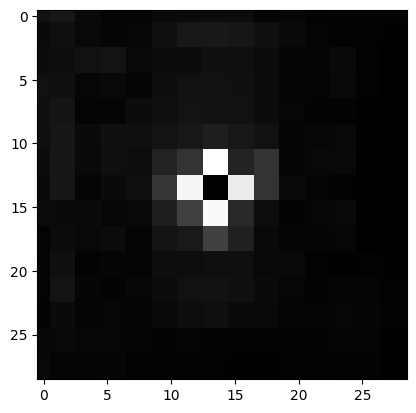}}
\subfigure[]{\includegraphics[width=0.2\linewidth]{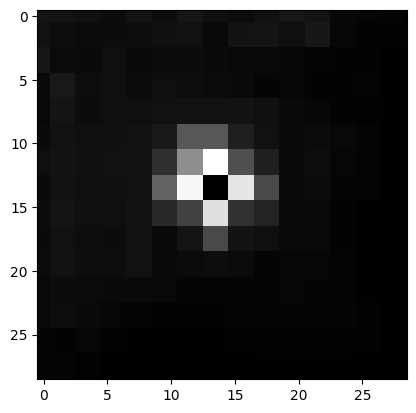}}
\subfigure[]{\includegraphics[width=0.2\linewidth]{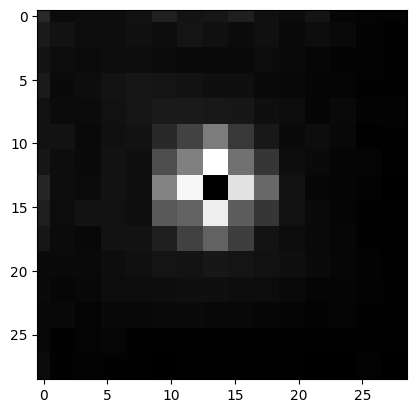}}
\subfigure[]{\includegraphics[width=0.2\linewidth]{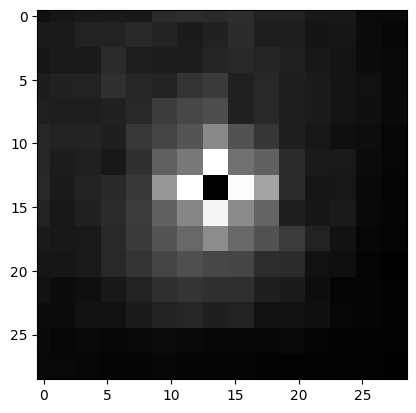}}
\caption{ 
We demonstrate the sensitivity of the Action Bisimulation encoder to changes in obstacles around the agent as we change the value of $c$ from left to right with $0.25$, $0.75$, $0.85$, $0.99$. 
}
\vskip -0.5cm
\label{fig:gamma_variance}
\end{figure}

Figure~\ref{fig:gamma_variance} shows how this dropoff varies as the value of $c$, the discount factor in Equation~\ref{actionbisimulation}, changes. The multi-step encoder gracefully increases in sensitivity with greater $c$, though a very large $c$ can make it unstable. Fundamentally, the possible sequences of actions grows exponentially, especially when trained with random actions, which is why selecting a value of c which ensures some dropoff ensures that the action-bisimulation representation does not become too off-policy.

\begin{figure}[H]
\centering     
\subfigure[]{\includegraphics[width=0.37\linewidth]{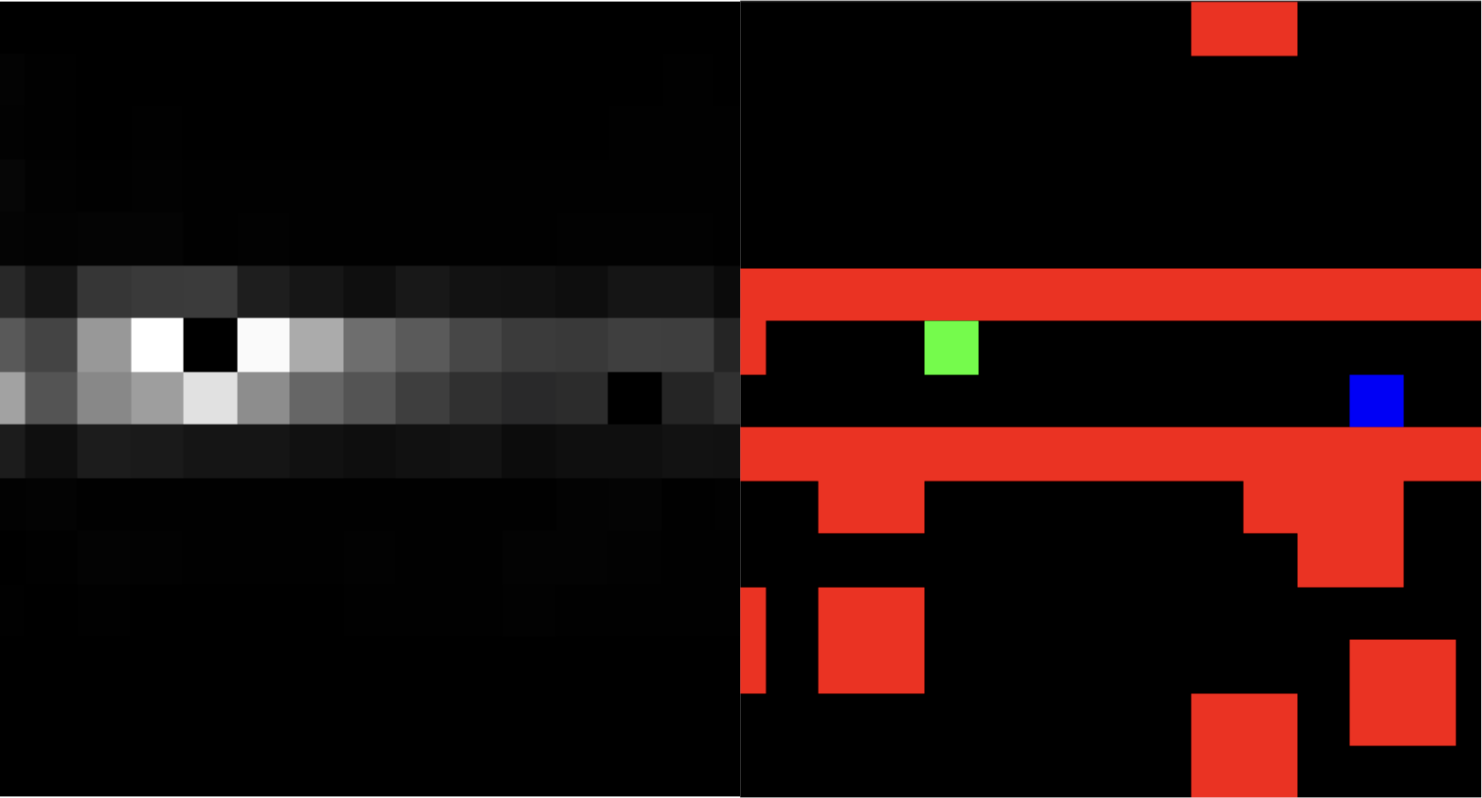}}
\subfigure[]{\includegraphics[width=0.4\linewidth]{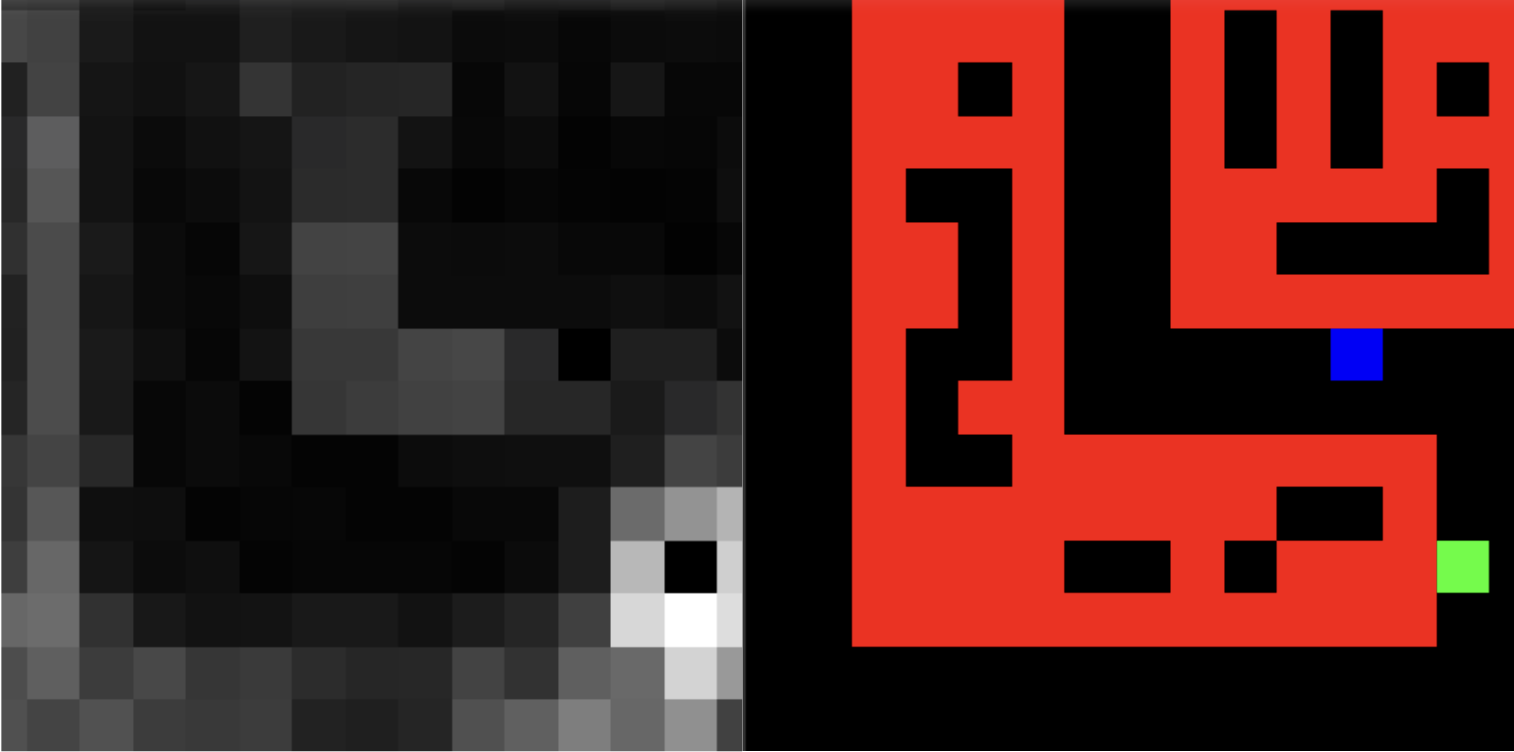}}
\caption{ \textbf{Left: Nav2D Corridor} shows how the action-bisimulation representation is sensitive to changes only within the corridor and agnostic to those without. Sensitivity is denoted by brighter pixels in the left subfigure. \textbf{Right: Nav2D Maze} demonstrates the action-bisimulation representation's sensitivity in a maze environment.
}
\vskip -0.5cm
\label{fig:maze_corridor}
\end{figure}

In Figure \ref{fig:maze_corridor}, we demonstrate how the action-bisimulation metric can learn reprsentations that ignore control-irrelevant information. In the Corridor environment, the agent is never able to leave the interior of the corridor but can always observe the obstacles on the exterior. We see that the representation is sensitive only to changes within the corridor. These results are echoed in the more complex Maze environment where the unreachable obstacles inside the maze's walls have little to no effect on the agent's representation. In fact, we can see that the representation's region of sensitivity almost exactly matches the agent's reachable locations.

\section{Additional RL Results}
\label{alternative_training}

The Pointmaze environment which we evaluated with used a set of discrete actions. In general, evaluating $U(\mathcal A)$, the uniform distribution over actions, is easier with continuous actions. Action-bisimulation can be approximated in continuous contexts simply by sampling some representative number of states. We demonstrate this in a continuous pointmaze environment, where the agent takes continuous 2D directions.

\begin{figure}[H]
\centering     
\includegraphics[width=0.5\linewidth]{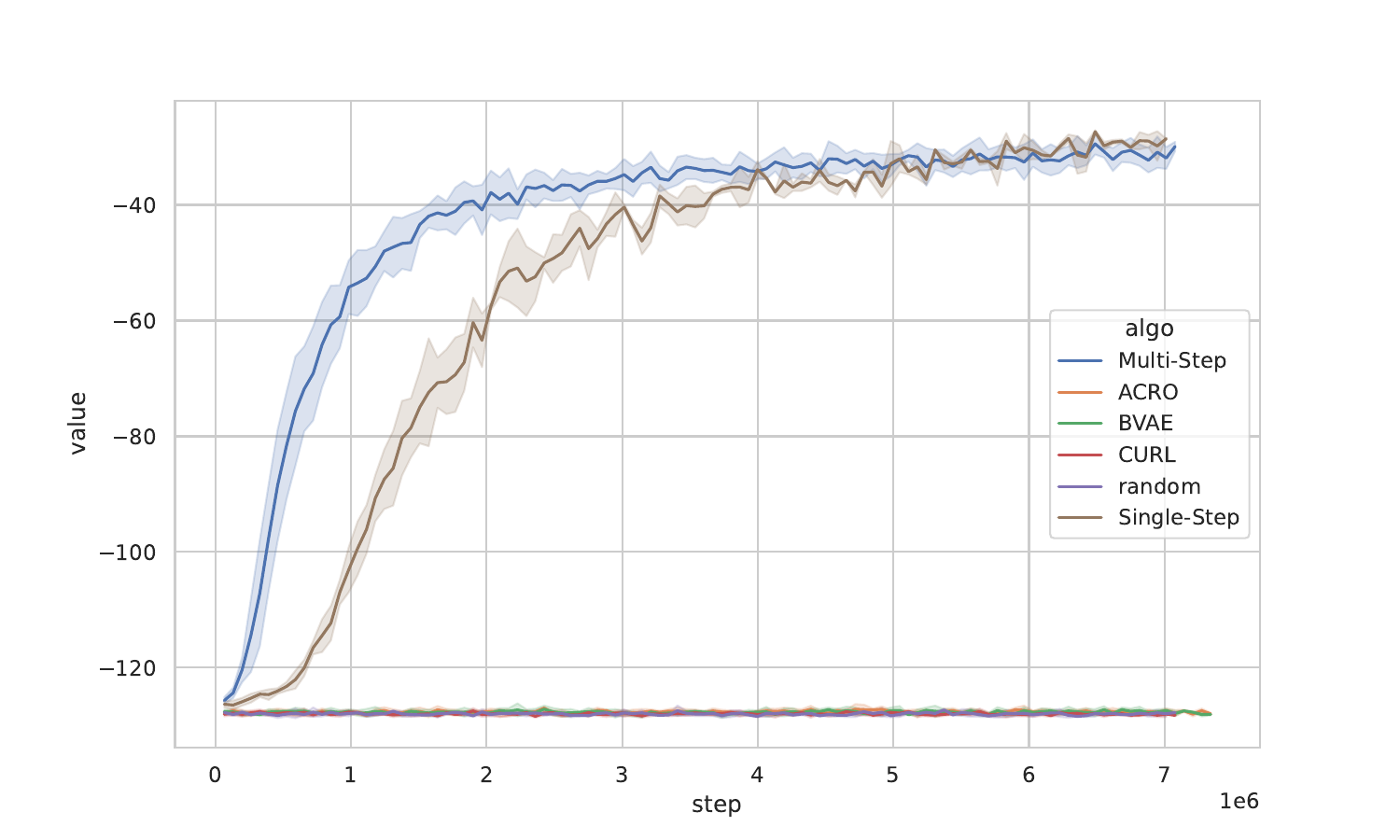}
\caption{
\textbf{Continuous Pointmaze performance}: because the action space is more challenging, many of the baselines struggle, especially ACRO, where action ambiguity is heightened.
}
\vskip -0.5cm
\label{fig:flow_ihrl}
\end{figure}

As we can see, Figure \ref{fig:flow_ihrl} provides evidence that action-bisimulation encodings are not limited to discrete actions.


\begin{table}[H]
\setlength\tabcolsep{3pt}
\begin{center}
\begin{small}
\begin{tabular}{|l||c|c|c|r|}
\hline
Environment & 2D Navigation & Point-Mass & Habitat \\\hline
Action-Bisimulation & $-14.266\pm0.509$ &$-30.8\pm3.7$ & $0.7754\pm0.005049$\\
Single-step & $-12.613\pm1.992$ &$-29.7\pm3.9$ & $0.7716\pm0.1629$\\
ACRO & $-49.436\pm0.342$ & $-127.9\pm0.61$ & $0.7374\pm0.0065$\\
$\beta$-VAE & $-13.880\pm2.263$ & $-128.4\pm0.52$ & $0.138\pm0.0220$\\
CURL & $-44.791\pm7.472$ & $-128.4\pm0.45$ & $0.7657\pm0.0193$\\
Vanilla & $-14.523\pm1.560$ & $-128.4\pm0.052$ & $0.728\pm0.0294$\\
\hline
\end{tabular}
\caption{Final Performance Evaluation}
\vskip -0.6cm
\label{finalperformancetable}
\end{small}
\end{center}
\end{table}
\section{Limitations}
\label{sec:limitations}
The three primary limitations we observed in this work for implementing action-bisimulation are as follows, and we go into further detail in the subsequent subsections:
\begin{enumerate}
    \item The minimum controllable single-step representation, especially when using learned inverse dynamics can omit controllable information if the action is overrepresented in the state 
    \item Uncontrollable, but reward-relevant elements must be incorporated into the representation after it is learned. 
    \item Both the forward model for transitions and the action-bisimulation encoding representation are bootstrapped over the expectation over all actions. This can result in unstable training.
    \item When a task does not require much lookahead, action-bisimulation will only provide a marginal benefit.
\end{enumerate}

\subsection{Limitations of Minimum Controllable Representation}
While~\cref{theory} demonstrates that at convergence the action-bisimulation encoding will capture only action-relevant information, it does not guarantee that it will capture \textit{all} action relevant information. If using the regularized single step loss Equation~\ref{RegularizedInverseDynamics}, the method is regularized to capture the \textit{minimal sufficient information} to predict actions. In practice, this can be quite limiting.

For example, consider the scenario where in the top corner of the screen there is a small display of the last action that the agent takes. In this scenario, the inverse dynamics model is likely to learn to only pay attention to this part of the screen, ignoring other components such as the state of the agent. This is because paying attention to this part is a sufficient representation of actions, even though it does not capture all action-relevant information. Information-based methods such as~\cref{ActioninfoNCE} or generative controllability representations are a possible solution for this, but we have found empirically that they do not seem to learn representations useful for RL. As a result, future work is for investigating action-controllable components that are not necessary for inverse dynamics (action prediction).

\subsection{Uncontrollable Reward-relevant components}
Another possible limitation is relevant for all controllability pertaining: a representation that captures controllable elements may fail to capture uncontrollable reward-relevant components. For example, consider a goal-based task where the goal is part of the state. The goal itself is not controllable and thus the representation of the goal will not be encoded in an action-bisimulation representation. While the action-bisimulation method is task-agnostic, at least insofar as the initial offline dataset is task-agnostic, it is not a sufficient representation in every task.

This issue can be mitigated by a variety of approaches. The simplest is to simply allow the representation to be modified to be task-specific in the RL training, and we employ this strategy in this work. However, more complex strategies might add an additional channel for task-relevant information, or integrating classic reward-bisimulation to learn the task-relevant components on top of the pre-trained action-bisimulation ones.

\subsection{Training Instability of Bootsrapping}
One of the challenges when learning an action-bisimulation representation is the inherent bootstrapping where the forward model is trained with $f(\phi(s), a) \rightarrow \phi'(s)$, and the encodings themselves are being updated with Equation~\ref{bisim_loss}. Since the action distribution in Equation~\ref{actionbisimulationlearn} is over the uniform expectation over actions, this can result in instability because of the combinatorial complexity of actions. One way we mitigated this is through the adaptive learning rate of the forward model, but future work should investigate stabilizing the convergence, especially if action-bisimulation is applied to online data.

\subsection{Tasks without Lookahead}
Finally, while multi-step controllability is a powerful property, not all tasks require this kind of lookahead, and it is not clear that multi-step pretraining would outperform single-step or other baselines in these cases. For example, in the popular Mujoco locomotion domains~\citep{ todorov2012mujoco}, knowing about future control can often be distracting to the agent---all the relevant information is captured by determining how the current action will affect future actions. Domains where long-term control is useful, such as manipulation, can also be challenging for the current form of action-bisimulation because of the minimal sufficient information property of the single-step losses. Future work is aimed at investigating this in greater detail.

\section{Baseline Details}
\label{baseline_details}
A detailed description of each of the baselines and the limitations of each. Also a mention of reward-based bisimulation.
\section{Environment Details}
\label{environment_appendix}
\begin{table}[H]
\setlength\tabcolsep{3pt}
\begin{center}
\begin{small}
\begin{tabular}{|l||c|c|r|}
\hline
Environment & Pretrain dataset & Evaluation Steps \\\hline
2D Navigation & $1$M samples &$2$M steps\\
Point-Mass & $0.25$M samples &$7$M steps\\
Habitat & $100$k samples & $2$M steps\\
\hline
\end{tabular}
\caption{Amount of data used for 2 phases. The pretrain dataset uses random actions and is used to train the encoders, and the evaluation steps is the number of environment steps used to train RL. }
\vskip -0.8cm
\label{DatasetTable}
\end{small}
\end{center}
\end{table}

\subsection{2D Navigation with Obstacles}
This environment is a 2D gridworld which consists of a $15\times15$ grid. The agent has 4 actions, up, down, left and right $[(0,-1), (0,1),(-1,0)(1,0)]$. If the agent moves into an obstacle, or the edge of the screen, its location will not change, otherwise, the direction will be added to the current position. and takes as observation a 3-channel $15\times 15$ image. The first channel encodes the agent position with $1$ at the location of the agent, and $-1$ elsewhere. The second channel encodes obstacles as $1$ where there is an obstacle and $-1$ otherwise, and the last channel encodes the goal. In this version, the goal is always located at the center of the image $(7,7)$. This is because otherwise a task-agnostic encoding would have to re-learn the goal location. The agent receives a reward of $-1$ everywhere except the goal, where it receives reward of $0$.

Initialization of the environment is as follows: the agent is initialized at a random location. Then obstacles are generated as $20$ $2\times2$ obstacles, initialized at random locations. The obstacles can be overlapping, but they cannot be initialized on top of the agent. Finally, the environment checks that there exists a path from the agent to the goal. if there is not, the environment is reinitialized until there is. Each episode is $50$ timesteps, after which a new environment is initialized.

\subsection{Pointmass}
This environment is a modification of the Mujoco Pointmass environment, where a pointmass with the dynamics of a damped linear $x$ and $y$ joint with damping coefficent $1$ and friction coefficient $0.5$ with navigates through the environment taking a set of four discrete actions, up, down, left, and right. The original environment only included a small number of pre-defined mazes in a $15\text{m}\times15\text{m}$ world. Additionally the original environment directly gives observations of the position of the goal and agent, while this version gives pixel data from a fixed topdown camera. This environment lacked the complexity of controllability in the dynamics we are interested in investigating in this work. Instead, we modified the environment so that $20$ $2\text{m}\times2\text{m}$ obstacles are randomly arranged in the environment, and added walls to prevent the point from leaving the field of view. The goal is always located in the center of the image. In this environment, the extrinsic reward function is a sparse 0/1 reward for being within $1m$ of the goal, which is the distance traversed by the agent in $1$ timestep. The agent takes episodes of $128$ time steps.\\

\subsection{Habitat}
Habitat~\citep{habitat19iccv} is a photorealistic 3D simulator for training embodied agents. The experiments in this paper use five scenes from the \texttt{Tiny} partition of the Gibson dataset~\citep{xiazamirhe2018gibsonenv}, \texttt{Andover}, \texttt{Azusa}, \texttt{Anaheim}, \texttt{Ballou},  and \texttt{Spotswood}. These scenes were chosen for their high navigational complexity. The observation space is a visually rich RGB+Depth image. Unlike the original Habitat environment, we choose to use an orthographic (as opposed to pinhole) camera placed above the goal in each episode so that the goal location is always at the center of the image observation; using a consistent goal location with respect to the camera is critical as we do not include any other goal information in the observation (in contrast with the traditional PointNav task in Habitat that includes a distance+compass heading sensor to the goal). In the RGB observation, we place a yellow box on top of the agent to indicate its location because the default rendered agent is sometimes the same color as the floor below it; the depth image remains unchanged. The agent and goal are spawned in new locations every episode such that the agent is always in view of the camera; this means that each episode looks at a different part of the scene. 
\section{Hyperparameters}
\label{hyperparameter_appendix}
\subsection{Nav2D}
The network dimensions and architectures used for Nav2D.

\begin{table}[H]
\centering

\textbf{Inverse Dynamics Model}\\
\begin{tabular}{ |p{3cm}||p{3cm}|  }
 \hline
 Layer Type & Layer Size \\
 \hline
 Input          & 1152  \\
 Linear \& ReLU & 256 \\
 Linear         & 256  \\
 \hline
\end{tabular}

\vspace{0.2cm}

\textbf{Action-Bisimulation Parameters}\\
\begin{tabular}{ |p{4cm}||p{2cm}|  }
 \hline
 Parameter & Value \\
 \hline
 Single Step \(L_1\) Penalty & \(0.0001\)  \\
 Multi Step \(c\) & \(0.99\) \\
 Learning Rate & \(0.0001\) \\
 \hline
\end{tabular}


\textbf{Reinforcement Learning Parameters}\\
\begin{tabular}{ |p{4cm}||p{2cm}|  }
 \hline
 Parameter & Value \\
 \hline
 Algorithm & DQN \\ 
 Batch Size \(L_1\) Penalty & \(32\)  \\
 $\epsilon_\text{end}$ & 0.2 \\
 $\epsilon_\text{start}$ & 0.9 \\
 $\gamma$ & \(0.99\) \\
 Learning Rate & 0.0001 \\
 \hline
\end{tabular}
\end{table}

\subsection{Pointmass}
The network dimensions and architectures used for the environment.

\vspace{0.2cm}

\begin{table}[H]
\centering
\textbf{Encoder parameters} \\
\begin{tabular}{ |p{3cm}||p{3cm}|p{3cm}|  }
 \hline
 Layer Type & Layer Size & Kernel Size \\
 \hline
 Input          & N/A & 64x64x3  \\
 Conv2D \& ReLU & 3x3 & 32x32x8  \\
 Conv2D \& ReLU & 3x3 & 16x16x16 \\
 Conv2D         & 8x8 & 1x1x32   \\
 \hline
\end{tabular}

\textbf{Inverse Dynamics Model}\\
\begin{tabular}{ |p{3cm}||p{3cm}|  }
 \hline
 Layer Type & Layer Size \\
 \hline
 Input          & 64  \\
 Linear \& ReLU & 256 \\
 Linear         & 32  \\
 \hline
\end{tabular}

\vspace{0.2cm}

\textbf{Actor/Critic Models}\\
\begin{tabular}{ |p{3cm}||p{3cm}|  }
 \hline
 Layer Type & Layer Size \\
 \hline
 Input          & 32  \\
 Linear \& ReLU & 256 \\
 Linear         & 4/1   \\
 \hline
\end{tabular}

\vspace{0.2cm}

\textbf{Action-Bisimulation Parameters}\\
\begin{tabular}{ |p{4cm}||p{2cm}|  }
 \hline
 Parameter & Value \\
 \hline
 Single Step \(L_1\) Penalty & \(1.0\)  \\
 Multi Step \(c\) & \(0.75\) \\
 K Steps         & \(5\)   \\
 Learning Rate & \(0.0001\) \\
 \hline
\end{tabular}

\vspace{0.2cm}

\textbf{Reinforcement Learning Parameters}\\
\begin{tabular}{ |p{4cm}||p{2cm}|  }
 \hline
 Parameter & Value \\
 \hline
 Algorithm & PPO \\ 
 Batch Size \(L_1\) Penalty & \(256\)  \\
 Steps Per Rollout & \(65536\) \\
 Steps Per Eval         & \(16384\)   \\
 Learning Rate & \(0.000025\) \\
 \hline
\end{tabular}
\end{table}

The \(L_1\) penalty is a particularly sensitive parameter, with this incorrectly set the single step model fails to identify relevant features. To train this effectively an adaptive term was used to scale the \(L_1\) regularization term to the listed value as the encoder approached convergence.

\subsection{Habitat}

\textbf{Encoder hyperparameters and PPO} 
The network dimensions and architectures used for the Habitat experiments are exact copies of the ResNet18~\citep{he2015deep} networks used in the original Habitat PointGoal navigation task~\citep{habitat19iccv}. For pretraining the encoders, we only trained the visual features encoder of the ResNet18 policy used in Habitat. Further, we used the vanilla implementation of PPO written in Habitat with all default parameters. 

\begin{table}[H]
\centering

\textbf{Inverse Dynamics Model}\\
\begin{tabular}{ |p{3cm}||p{3cm}|  }
 \hline
 Layer Type & Layer Size \\
 \hline
 Input          & 2048  \\
 Linear \& ReLU & 256 \\
 Linear         & 256  \\
 \hline
\end{tabular}

\vspace{0.2cm}

\textbf{Action-Bisimulation Parameters}\\
\begin{tabular}{ |p{4cm}||p{2cm}|  }
 \hline
 Parameter & Value \\
 \hline
 Single Step \(L_1\) Penalty & \(0.0\)  \\
 Multi Step \(c\) & \(0.95\) \\
 Learning Rate & \(0.0001\) \\
 \hline
\end{tabular}

\vspace{0.2cm}

\textbf{PPO Parameters}\\
\centering
\begin{tabular}{|p{4cm}||p{2cm}|}
\hline
\textbf{Parameter} & \textbf{Value} \\
\hline
clip\_param       & 0.2      \\
ppo\_epoch        & 4        \\
num\_mini\_batch  & 2        \\
value\_loss\_coef & 0.5      \\
entropy\_coef     & 0.01     \\
lr                & 0.00025   \\
eps               & .00001     \\
max\_grad\_norm   & 0.5      \\
num\_steps        & 128      \\
hidden\_size      & 512      \\
gamma             & 0.99     \\
tau               & 0.95     \\
\hline
\end{tabular}
\label{tab:hyperparameters}
\end{table}

\end{document}